%% file: main.tex
\documentclass{article}

\usepackage[margin=1in]{geometry}
\usepackage{mathpazo}
\usepackage{booktabs}

\usepackage{amsmath,amssymb,amsthm}
\usepackage{xspace}
\usepackage{graphicx}
\usepackage{color}
\usepackage[colorlinks=true,linkcolor=blue,citecolor=blue,urlcolor=blue]{hyperref}
\usepackage{todonotes}
\usepackage{subcaption}
\usepackage{comment}
\usepackage[ruled]{algorithm2e}
\usepackage[T1]{fontenc}
\usepackage[backend=biber,style=alphabetic,natbib=true,maxbibnames=99,minalphanames=3]{biblatex}
\newcommand{\N}{\mathbb{N}}
\newcommand{\Z}{\mathbb{Z}}

\newcommand{\E}[2]{\mathbb{E}_{{#1}}\left[{#2}\right]}

\newcommand{\codelink}{\xspace{\url{https://git.io/unadversarial}}}

\title{\Large Unadversarial Examples: Designing Objects for Robust Vision}
\author{
    Hadi Salman\footnote{Equal contribution.} \\
    \texttt{hadi.salman@microsoft.com} \\
    Microsoft Research
    \and
    Andrew Ilyas\footnotemark[1] \\
    \texttt{ailyas@mit.edu} \\
    MIT
    \and 
    Logan Engstrom\footnotemark[1] \\
    \texttt{engstrom@mit.edu} \\
    MIT
    \and
    Sai Vemprala \\
    \texttt{saihv@microsoft.com} \\
    Microsoft Research
    \and
    Aleksander M\k{a}dry \\
    \texttt{madry@mit.edu} \\
    MIT
    \and 
    Ashish Kapoor \\
    \texttt{akapoor@microsoft.com} \\
    Microsoft Research
}

\date{}
\begin{document}
    \maketitle
    \begin{abstract}
        \input{abstract}
    \end{abstract}

    \section{Introduction}
    \label{sec:intro}
    \input{sections/intro}

    \section{Motivation and Approach}
    \label{sec:motivation}
    \input{sections/motivation}

    \subsection{Constructing unadversarial objects}
    \label{sec:algorithms}
    \input{sections/methods}

    \section{Experimental Evaluation}
    \label{sec:experiments}
    \input{sections/experiments}

    \section{Related Work}
    \label{sec:related}
    \input{sections/related}

    \section{Discussion and Conclusions}
    \label{sec:conclusions}
    \input{sections/conclusions}

    \section*{Acknowledgements}
    We are grateful to Ian Engstrom for helping take the photographs for the
    physical-world experiments.

    Work supported in part by the NSF grants CCF-1553428 and CNS-1815221, and
    the Microsoft Corporation. This material is based upon work supported by the
    Defense Advanced Research Projects Agency (DARPA) under Contract No.
    HR001120C0015.

    Research was sponsored by the United States Air Force Research Laboratory and
    was accomplished under Cooperative Agreement Number FA8750-19-2-1000. The views
    and conclusions contained in this document are those of the authors and should
    not be interpreted as representing the official policies, either expressed or
    implied, of the United States Air Force or the U.S. Government. The U.S.
    Government is authorized to reproduce and distribute reprints for Government
    purposes notwithstanding any copyright notation herein. 
    \clearpage

    \printbibliography
    \clearpage
    \input{appendix/appendix}

    \clearpage

    \end{document}

%% file: abstract.tex
We study a class of realistic computer vision settings wherein one can
influence the design of the objects being recognized. We develop a framework
that leverages this capability to significantly improve vision models’
performance and robustness. This framework exploits the sensitivity of modern
machine learning algorithms to input perturbations in order to design
``robust objects,’’ i.e., objects that are explicitly optimized to be
confidently detected or classified. We demonstrate the efficacy of the
framework on a wide variety of vision-based tasks ranging from standard
benchmarks, to (in-simulation) robotics, to real-world experiments. Our
code can be found at \codelink.

%% file: sections/intro.tex
Performing reliably on unseen or shifting data distributions is a difficult
challenge for modern computer vision systems.
For example, slight rotations and translations of images suffice to reduce the
accuracy of state-of-the-art
classifiers~\citep{engstrom2019rotation,alcorn2019strike,kanbak2018geometric}. 
Similarly, models that attain near human-level performance on benchmarks 
exhibit significantly degraded performance when faced with even mild image corruptions and
transformations~\citep{hendrycks2019benchmarking, kang2019testing}.
Furthermore, when an adversary is allowed to modify inputs directly, standard vision models
can be manipulated into predicting arbitrary outputs (cf. {\em
adversarial examples} \citep{biggio2013evasion, szegedy2014intriguing}).
While robustness interventions and additional training data 
can improve out-of-distribution behavior, they do not fully
close the gap between model performance on standard heldout data and on
corrupted/otherwise unfamiliar data~\citep{taori2020when,
hendrycks2020faces}. The situation is worse still when 
test-time distribution is under- or mis-specified, which occurs commonly in
practice. 

How can we change this state of affairs? 
We propose a new approach to image recognition in the face of unforeseen corruptions or
distribution shifts. This approach is rooted in a reconsideration of the problem setup itself.
Specifically, we observe that in many situations, a system designer not only
trains the model used to make predictions, but also controls, to some degree,
the inputs that are fed into that model.
For example, a drone operator seeking to train a landing pad detector can modify the
surface of the landing pad; and, a roboticist training a perception model to recognize a
small set of custom objects can slightly alter the texture or design of
these objects. 

\begin{figure*}[!h]
    \centering
    \includegraphics[width=0.9\linewidth]{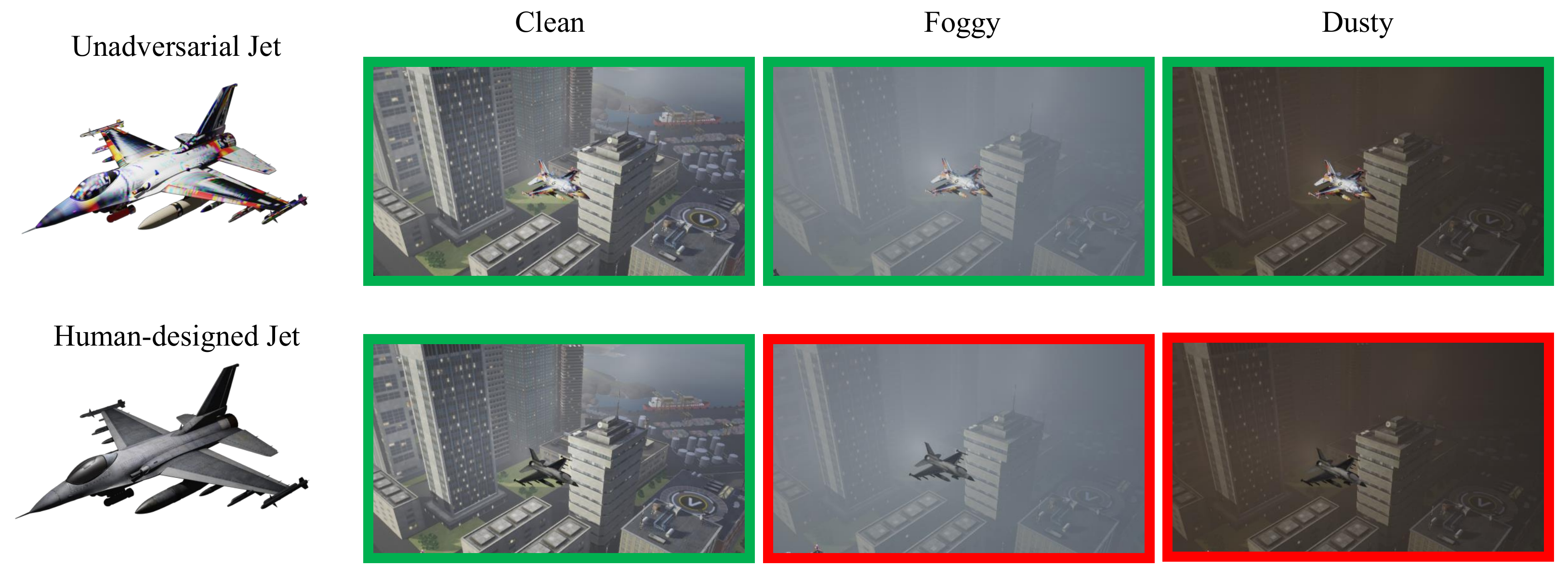}
    \caption{In this paper, we demonstrate that optimizing objects (e.g., the
    pictured jet) for pre-trained neural networks can significantly boost
    performance and robustness on computer vision tasks.}
    \label{fig:headline}
\end{figure*}

We find that such control over inputs can be leveraged to drastically improve our ability to tackle computer vision tasks. In particular, it allows us to turn the
input-sensitivity of modern vision systems from a weakness into a strength.
That is, instead of optimizing inputs to {\em mislead} models (as in
adversarial examples), we can alter inputs to {\em reinforce} correct behavior,
yielding what we refer to as ``unadversarial examples.''
Indeed, we show that even a simple gradient-based algorithm can successfully
construct such unadversarial
examples for a variety of vision settings and demonstrate that, by optimizing objects for vision
systems (rather than only vice-versa), we can significantly improve both
in-distribution performance \emph{and} robustness to unforeseen data shifts and corruptions.

Specifically, in this paper, we develop methods for  constructing unadversarial stickers/patches that boost the performance of the deep learning-based image classifiers operating on the corresponding objects.  We then demonstrate the efficacy
    of these methods on both standard benchmarks (CIFAR, ImageNet) and
    robustness-based benchmarks (ImageNet-C, CIFAR-C) while also comparing them to a broad set of baselines, including QR codes. To further highlight the practicality of our framework, we (a) extend
    our methods to designing the texture of three-dimensional objects
    (rather than patches); (b) deploy unadversarial examples in a
    simulated drone setting; and (c) ensure that the performance improvement yielded by the objects we design
    actually transfer to the physical world.

%% file: sections/motivation.tex
While vision models tend to perform well on held-out data drawn from the same
distribution as the training data, out-of-distribution inputs can severely degrade this
performance. 
For example, models behave unreliably under
distribution shifts induced by new data collection
procedures~\citep{recht2018imagenet,engstrom2020identifying,torralba2011unbiased},
synthetic corruptions~\citep{hendrycks2019benchmarking,kang2019testing}, 
spatial transformations~\citep{engstrom2019rotation,fawzi2015manitest}, as well
as under other types of shift.

Given a fixed type of distribution shift, a standard approach to increasing
model robustness is to explicitly train on or regularize with data from the corresponding
anticipated test distribution \citep{kang2019testing}. For example,
\citet{engstrom2019rotation} find that 
vision models trained on worst-case rotations and
translations end up being fairly robust to rotation and translation-based
distribution shifts. 
However, this approach is not without shortcomings---for example, \citet{kang2019testing}
find that 
training CIFAR classification models that are robust to 
JPEG-compression in this manner requires a significant sacrifice in natural accuracy. 
Recent works make similar observations in the context of other
distribution shift mechanisms like $\ell_p$
adversaries~\citep{tsipras2019robustness,su2018robustness, 
raghunathan2019adversarial} or texture swapping
\citep{geirhos2018imagenettrained}. 

These observations give rise to a more general question:
given that performing reliably in the face of constrained, well-specified
distribution shifts is already a difficult challenge, 
how can we attain robustness to broad, unforeseen distribution shifts?

\subsection{Leveraging more controlled vision settings}
\label{subsec:controlled_settings_motivation}
Consider the vision tasks of detecting a landing pad from a drone, or
classifying manufacturing components from a factory robot. In both these tasks,
reliable in-distribution performance is a necessity; still,
a number of possible distribution shifts may occur at deployment time. For
example, the drone might approach the landing pad at an atypical angle, or
have a view obstructed by snow, smoke, or rain. Similarly, the factory robot may
encounter objects in unfamiliar poses, or using a low-quality/noisy camera. 

At first glance, dealing with these issues seems to require tackling the
difficult problem of general distribution shift robustness 
discussed earlier in this section.
However, there is in fact a critical distinction between the scenarios
considered above and vision tasks in their full generality. In particular,
in these scenarios and many others, the system designer has control over
not only the model that is used but also the physical objects that the model
operates on. As we will demonstrate, the designer can use this capability to
modify these objects to majorly boost  the model's ability to solve the problem at hand.

For instance, the designer of the drone's landing algorithm could, in
addition to training a detection model, also paint the landing pad bright
yellow. A machine learning model trained to detect this custom
landing pad might then be more effective than a model trained to detect a
standard grey pad, especially in low-visibility conditions.
Still, the particular choice to paint the landing pad yellow is rather ad hoc,
and likely rooted in the way {\em humans} recognize objects.  
Meanwhile, an abundance of prior work (e.g.,
\citep{jacobsen2019excessive, geirhos2018imagenettrained,
jetley2018friends, ilyas2019adversarial}) demonstrates that 
humans and machine learning models tend to use different sets of
features to make their decisions. This suggests that rather than relying on
human priors, we should instead be asking: {\em how can we build objects that are easily
detectable by machine learning models?}

\subsection{Unadversarial examples}
The task of making inputs {\em less} recognizable by computer vision systems has
been a focus of research in {\em adversarial examples}. Adversarial examples are
small, carefully constructed perturbations to natural images that can induce
arbitrary (mis)behavior from machine learning
models~\citep{biggio2013evasion,szegedy2014intriguing}. These perturbations
are typically constructed as the result of an optimization problem that
maximizes the loss of a machine learning model with respect to the input,
i.e., by solving the optimization problem
\begin{equation}
    \label{eq:adv_example}
    \delta_{adv} = \arg\max_{\delta \in \Delta} L(f_{\theta}(x + \delta), y),
\end{equation}
where $f_{\theta}$ is a parameterized model (e.g., a neural network with weights
$\theta$); $x$ is a natural input; $y$ is 
the corresponding correct label; $L$ is the loss function used to train $\theta$
(e.g., cross-entropy loss) and $\Delta$ is a
class of permissible perturbations (e.g., norm-bounded perturbations: $\Delta =
\{\delta: \|\delta\|_p \leq \epsilon\}$ for some small $\epsilon > 0$). 
Adversarial perturbations are typically crafted via projected gradient
descent (PGD) \citep{nesterov2003introductory} in input space,
a standard iterative first-order optimization method---prior work in adversarial examples has shown
that even a few iterations of PGD suffice to completely change the prediction of
many state-of-the-art machine learning systems \citep{madry2018towards}. 

\paragraph{From adversarial examples to unadversarial objects.}
The goal of this work is to modify the design of objects so that they are
more easily recognizable by computer vision systems. 
If we could specify every pixel of every image that a model encounters at
test time, we could
draw on the effectiveness of adversarial examples, and
construct image perturbations (using PGD) that {\em minimize} the loss of the
system, e.g.,
\begin{equation}
    \label{eq:unadv_example}
    \delta_{unadv} = \arg\min_{\delta \in \Delta} L(\theta; x + \delta, y).
\end{equation}
In our setting of interest, however, having such fine-grained access to the test
inputs is unrealistic (presumably, if we had precise control over every pixel in
the input, we could just directly encode the ground-truth label directly in it).
Instead, we have {\em limited} control over some physical objects; these objects
are in turn captured within image inputs, along with many signals that are
out of our control, such as camera artifacts, weather effects, or background
scenery.

It turns out that we can still draw on techniques from
adversarial examples research in this limited-control setting. 
Specifically, a recent line of work
\citep{kurakin2017adversarial, sharif2016accessorize, evtimov2018robust,
athalye2018synthesizing} concerns itself with constructing {\em robust
adversarial examples}~\citep{athalye2018synthesizing}, i.e., physically realizable
objects that act as adversarial examples when introduced into a scene in any one
of a variety of ways. For example, \citet{sharif2016accessorize} design
glasses frames that cause facial recognition models to misclassify faces,
\citet{athalye2018synthesizing} design custom-textured 3D models that are
misclassified by state-of-the-art ImageNet classifiers from many angles and
viewpoints, and~\citep{brown2018adversarial} design adversarial patches:
stickers that can be placed anywhere on objects causing them to be misclassified. 
In this paper, we leverage the techniques developed in the above line of work
to constructi robust un-adversarial objects---physically realizable {\em objects}
optimized to minimize (rather than maximize) the loss of a target
classifier. In the next section, we will more concretely discuss our methods for
generating unadversarial objects, then outline our evaluation setup.

%% file: sections/methods.tex
\begin{figure}
    \centering
    \begin{subfigure}{0.3\textwidth}
        \centering
        \includegraphics[width=\textwidth]{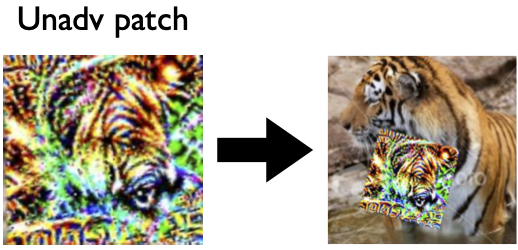}
        \caption{An example unadversarial patch designed for the ``tiger'' class.}
        \label{fig:example_unadv_patch}
    \end{subfigure}
    \hspace{2em}
    \begin{subfigure}{0.62\textwidth}
        \centering
        \includegraphics[width=\textwidth]{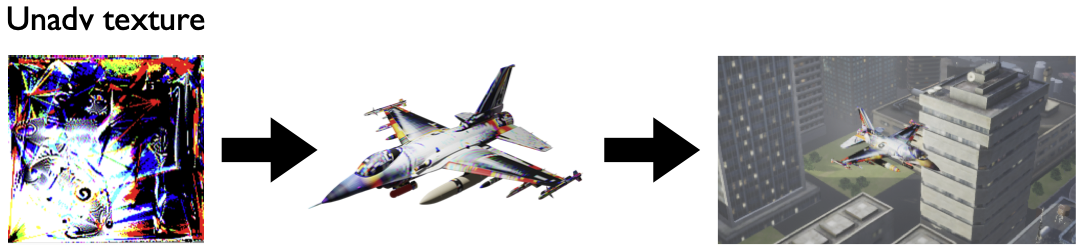}
        \caption{An example unadversarial texture designed for a jet 3D mesh (class
        ``warplane'') and applied to rendered city backgrounds.}
        \label{fig:example_unadv_texture}
    \end{subfigure}
    \caption{Examples of the two considered methods for constructing
    unadversarial objects.}
    \label{fig:example_unadvs}
\end{figure}

In the previous section, we identified a class of scenarios where a system
designer can not only control the machine learning model being deployed but
also, to some extent, the objects that model operates on. 
In these settings, we motivated so-called {\em unadversarial
examples} as a potential way to boost models' overall performance and robustness
to distribution shifts. 
In this section, we present and illustrate two concrete algorithms for
constructing unadversarial examples:
unadversarial patches and unadversarial textures. In the former, we
design a sticker or ``patch'' \citep{brown2018adversarial} that can be placed
on the object; in the latter, we design the 3D texture of the object (in a
similar manner to the texture-based adversarial examples of
\citet{athalye2018synthesizing}). Example results from both techniques are
shown in Figure~\ref{fig:example_unadvs}. For simplicity, we will assume that
the task being performed is image classification, but the techniques are
directly applicable to other tasks as well. In all cases, we require access to a
pre-trained model for the dataset of interest.

\smallskip
\noindent \textbf{Unadversarial patches (cf. Algorithm \ref{alg:patches} in Appendix \ref{app:algorithms}).} 
To train unadversarial patches (cf. Figure~\ref{fig:example_unadv_patch}), in addition to the pre-trained model, we require
sample access to image-label pairs from the dataset of interest. At each
iteration, we sample an image-label pair $(x, y)$ from a training set, and
place the patch corresponding to class $y$ onto the image with random
orientation and position\footnote{We allow the patch to be placed anywhere as
a matter of convenience: ideally we would only be applying the patch onto the
main object itself, but this would require bounding box data that we do not
have for most classification datasets.}. Since placing the patch is an affine
transformation, after each iteration we can compute the gradient of the
model's loss with respect to the pixels in the patch, and take a negative
gradient step on the patch parameters. The algorithm terminates when the
model's loss on sticker-boosted images plateaus, or after a fixed number of
iterations.

\smallskip
\noindent \textbf{Unadversarial textures (cf. Algorithm \ref{alg:textures} in Appendix \ref{app:algorithms}).} 
To train unadversarial {\em textures} (cf. Figure
\ref{fig:example_unadv_texture}), we do not require sample access to the
dataset, but instead
a set of 3D meshes for each class of objects that we would like to augment,
as well as a set of background images that we can use to simulate sampling a
scene (these can be images from the dataset of interest, solid-color
backgrounds, random patterns, etc.).

For each 3D mesh, our goal is to optimize a 2D texture which improves classifier
performance when mapped onto the mesh. 
At each iteration, we sample a mesh and a random background; we then use a 3D
renderer (Mitsuba \citep{mitsuba}) to map the corresponding texture onto the
mesh. We overlay the rendering onto a random background image, and then feed the
resulting composed image into the pre-trained classifier, with the label being
that of the sampled 3D mesh.
Since rendering is typically non-differentiable, we use a linear
approximation of the rendering process (cf. \citet{athalye2018synthesizing})
in order to compute (this time approximate) gradients of the model's loss
with respect to the utilized texture. From there, we apply the same SGD
algorithm as we did for the patch case.

%% file: sections/experiments.tex
In order to determine the effectiveness of our proposed framework, we evaluate
against a suite of computer vision tasks. We briefly outline the experimental
setup of each task below, and show that unadversarial objects
consistently improve the performance and robustness of the vision systems
tested. 
For a more detailed account of each experimental setup, see Appendix
\ref{app:exp_setup}; code for reproducing our experimental
results is available at \codelink.

\subsection{Clean data and synthetic corruptions}
\label{sec:eval_benchmark}
\begin{figure}[h]
    \centering
    \includegraphics[width=.45\textwidth]{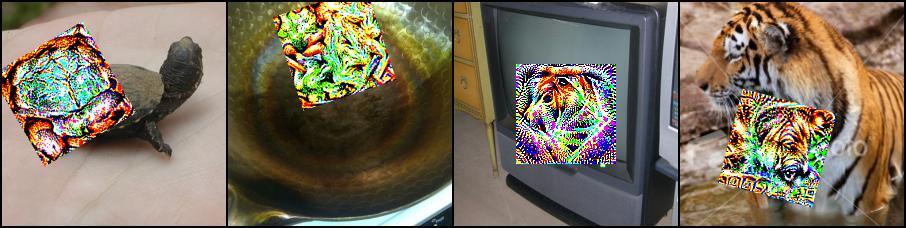}
    \quad\quad\quad
    \includegraphics[width=.45\textwidth]{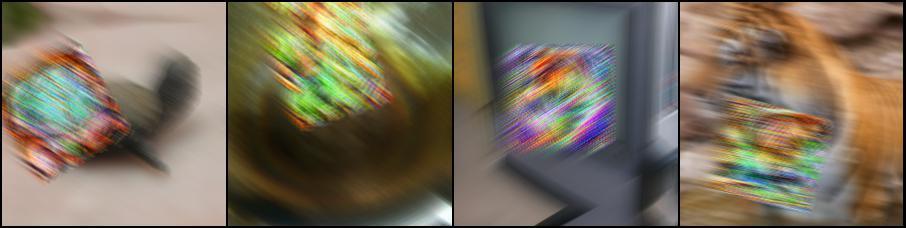}
    \caption{Clean (left) and corresponding corrupted (right)
    ImageNet images augmented with an unadversarial patch---we use such images
    to evaluate the efficacy of unadversarial patches in Section \ref{sec:eval_benchmark}.}
    \begin{subfigure}{0.29\textwidth}
        \includegraphics[width=\linewidth]{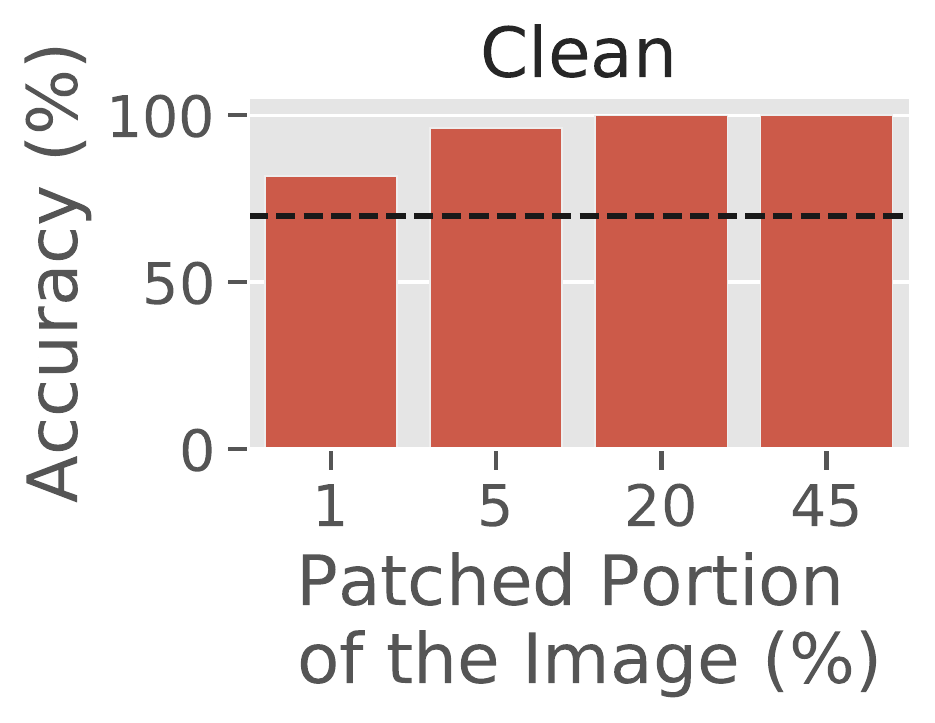}
        \caption{Performance on ImageNet}
        \label{fig:main_imagenet_clean}
    \end{subfigure}
    \begin{subfigure}{0.7\textwidth}
        \centering
        \includegraphics[width=\linewidth]{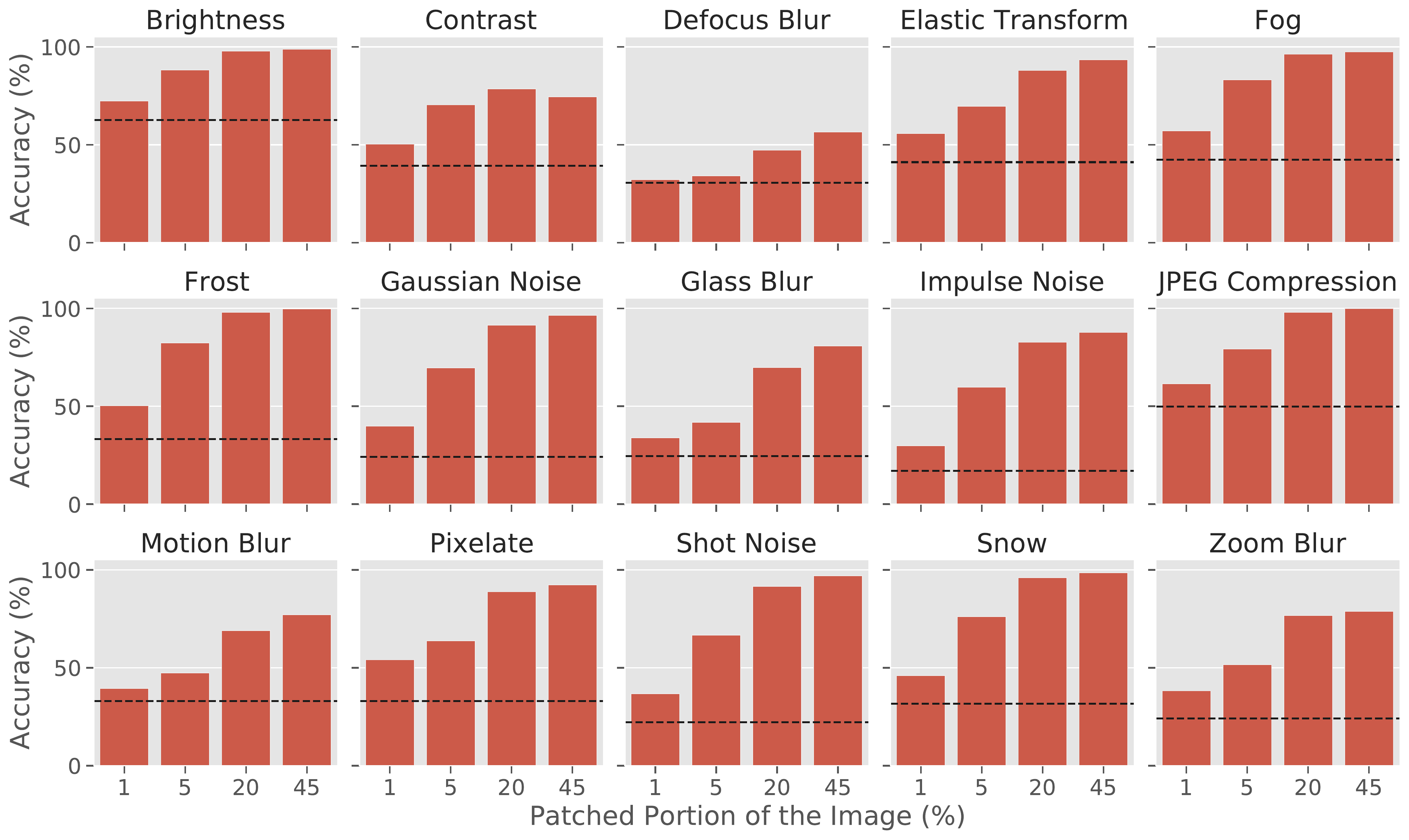}
        \caption{Performance on synthetically corrupted data (ImageNet-C)}
        \label{fig:2d_results}
    \end{subfigure}
    \caption{Accuracy on (a) clean ImageNet images and (b) synthetically
    corrupted ImageNet-C images as a function of patch size (given as a
    percentage of image area). In (b), each bar denotes the average accuracy
    over the five severities in ImageNet-C, and the horizontal dashed lines
    report the accuracy on the original (non-patched) datasets. 
    Unadversarial patches consistently boost performance for both clean and
    corrupted images, with accuracy monotonically increasing with patch size.
    The patches were trained without any corruptions or non-standard data
    augmentation in-the-loop (we train with the same augmentations that the
    pre-trained model itself was trained with).} 
\end{figure}

We first test whether unadversarial examples improve the
performance of image classifiers on benchmark datasets. 
Using the algorithm described in Section \ref{sec:algorithms}, we construct
unadversarial patches of varying size for pre-trained ResNet-50 classifiers on
the CIFAR \citep{krizhevsky2009learning} and ImageNet \citep{russakovsky2015imagenet}
datasets.
For evaluation, we add these patches at random positions, scales, and
orientations to validation set images (see Appendix~\ref{app:exp_setup} for the
exact protocol). As shown in Figure \ref{fig:main_imagenet_clean}, the pre-trained
ImageNet classifier is more consistently more accurate on the augmented ImageNet
images. For example, an  unadversarial patch 20 times smaller than ImageNet
images boosts accuracy by $26.3\%$ (analogous results for CIFAR are given in
Appendix \ref{app:detailed_results}). 

\paragraph{Robustness to synthetic corruptions.} 
Next, we use the CIFAR-C and ImageNet-C datasets 
\citep{hendrycks2019benchmarking} (consisting of the CIFAR and ImageNet
validation sets corrupted in 15 systematic ways) 
to see whether the addition of unadversarial patches to images confers any
robustness to image corruptions.  

We use the same patches and evaluation protocol that we used when looking at
clean data (to ensure a fair evaluation, we apply corruptions to boosted images
only {\em after} the unadversarial patches have been applied). As a consequence,
at test time neither model nor patch has been exposed to any image corruptions
beyond standard data augmentation. 
As a result, this experiment tests the ability for unadversarially
boosted images to withstand completely unforeseen corruptions; we also avoid any
potential biases from training on (and thus ``overfitting'' to
\cite{kang2019testing}) a specific type of corruption. 
The results (cf. Figure~\ref{fig:2d_results} for ImageNet and
Appendix~\ref{app:detailed_results} for CIFAR) indicate that unadversarial
patches do improve performance across corruption types; for example, applying an
unadversarial patch 5\% the size of a standard ImageNet image boosts accuracy by
an average of $31.7\%$ points across corruptions
\footnote{Since the original
corruption benchmarks proposed by \citep{hendrycks2019benchmarking} are only
available as pre-computed JPEGs (for which we cannot apply a patch
pre-corruption) or CPU-based Python image operations (which were
prohibitively slow), we re-implemented all 15 corruptions as batched GPU
operations; we verified that model accuracies on our corruptions mirrored the
original CPU counterparts (i.e., within $1\%$ accuracy). For more details about
our reimplementation, see our code release.}. 

\paragraph{Baselines.} We also compare our results to a variety
of natural baselines; the most relevant of these is the ``best loss image
patch,'' where we use the minimum-loss training image in place of a patch.
We compare with this baseline to ensure that our method is doing something
beyond this naive way to add signal to an image. The results are shown in
Appendix~\ref{app:detailed_results}, along with comparisons to less
sophisticated baselines, such as QR Codes and predefined random Gaussian noise patches.

\begin{figure*}[t]
    \centering
    \includegraphics[width=.9\linewidth]{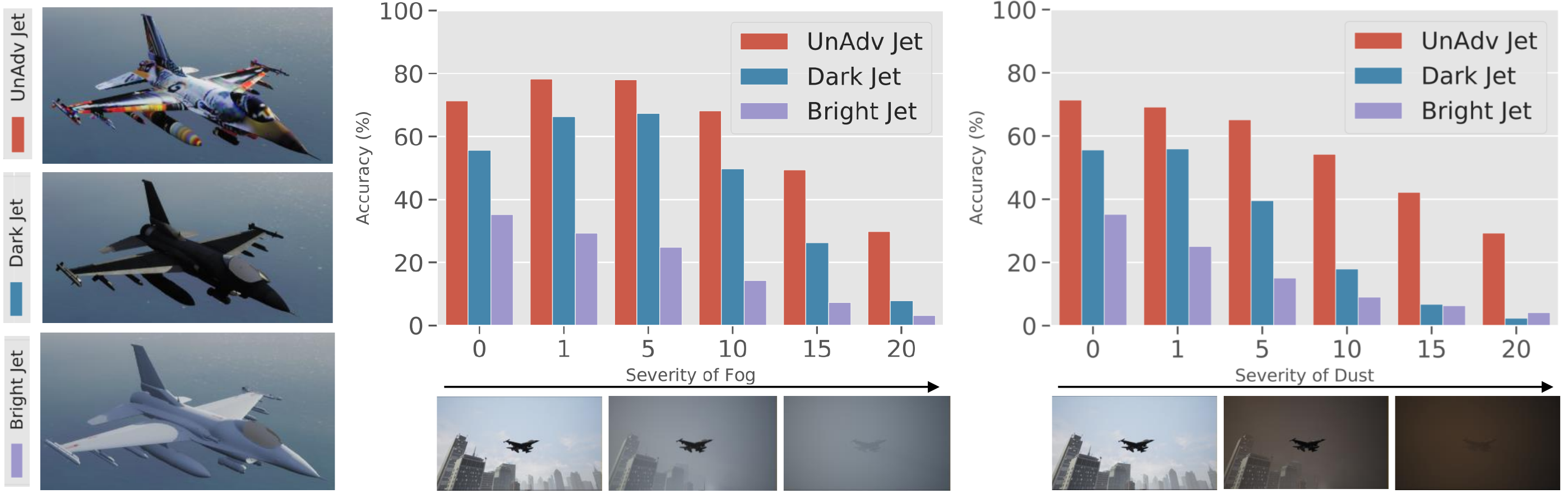}
    \caption{The jet unadversarial example task. We
    show example conditions under which we evaluate the objects, along with
    aggregate statistics for how well an ImageNet classifier classifies the objects in different conditions. We find that the classifiers
    perform consistently better on the unadversarial jet texture over the standard
    jet texture in both standard and distributionally shifted conditions.
    We also give a baseline of a white jet with a lighter texture 
    because of the poorly visibility inherent in the simulator; we find it performed 
    worse than even the standard jet.}
    \label{fig:jetfighter}
\end{figure*}
\begin{figure*}[t]
    \centering
    \includegraphics[width=1\linewidth]{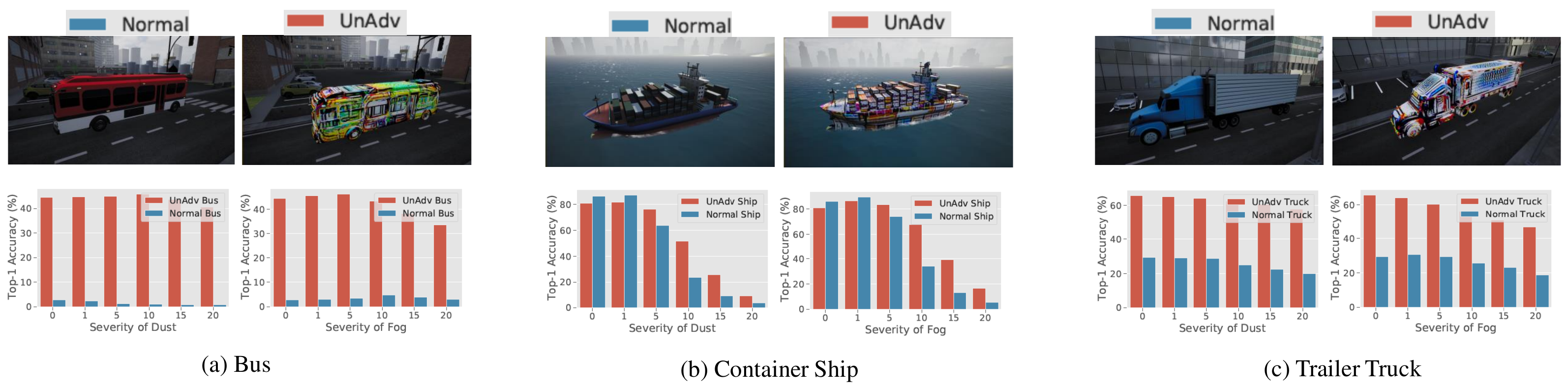}
    \caption{Additional examples reporting aggregate statistics for how well an ImageNet classifier classifies various objects in different conditions. Again, we find that the classifiers perform consistently better on the unadversarial objects texture over the standard objects.}
    \label{fig:other-3D-boosters}
\end{figure*}

\subsection{Classification in 3D simulation}
We now 
test unadversarial examples in a more practical setting:
recognizing three-dimensional objects in a high-fidelity similar.
We collect a set of three-dimensional meshes corresponding to four
ImageNet classes: ``warplane'', ``minibus'', ``container ship'', and ``trailer truck'' ,
sourced from \url{sketchfab.com}. We generate a texture
for each object using the unadversarial texture algorithm outlined in
Section~\ref{sec:algorithms}, using the ImageNet validation set
as the set of backgrounds for the algorithm, and a pre-trained ResNet-50 as the 
classifier.

To evaluate the resulting textures, we import each mesh into
Microsoft AirSim, a high-fidelity three-dimensional
simulator; we then test pre-trained ImageNet models' ability to recognize each
object with and without the unadversarial texture applied in a variety of
surroundings. We also test each 
texture's robustness to more realistic weather corruptions (snow and fog) built
directly into the simulator (rather than applied as a post-processing step). We
provide further detail on AirSim and our usage of it in Appendix~\ref{app:airsim}.
Examples of the images used to evaluate the unadversarial textures, as well as
our main results for one of the meshes are shown in Figure~\ref{fig:jetfighter}. 
We find that in both standard and simulated adverse weather conditions,
the model consistently performs better on the mesh wrapped in the unadversarial
texture than on the original. We present similar results for the other three
meshes in Figure~\ref{fig:other-3D-boosters}.

\subsection{Localization for (Simulated) Drone Landing}
\begin{figure*}
    \centering
    \includegraphics[width=.9\linewidth]{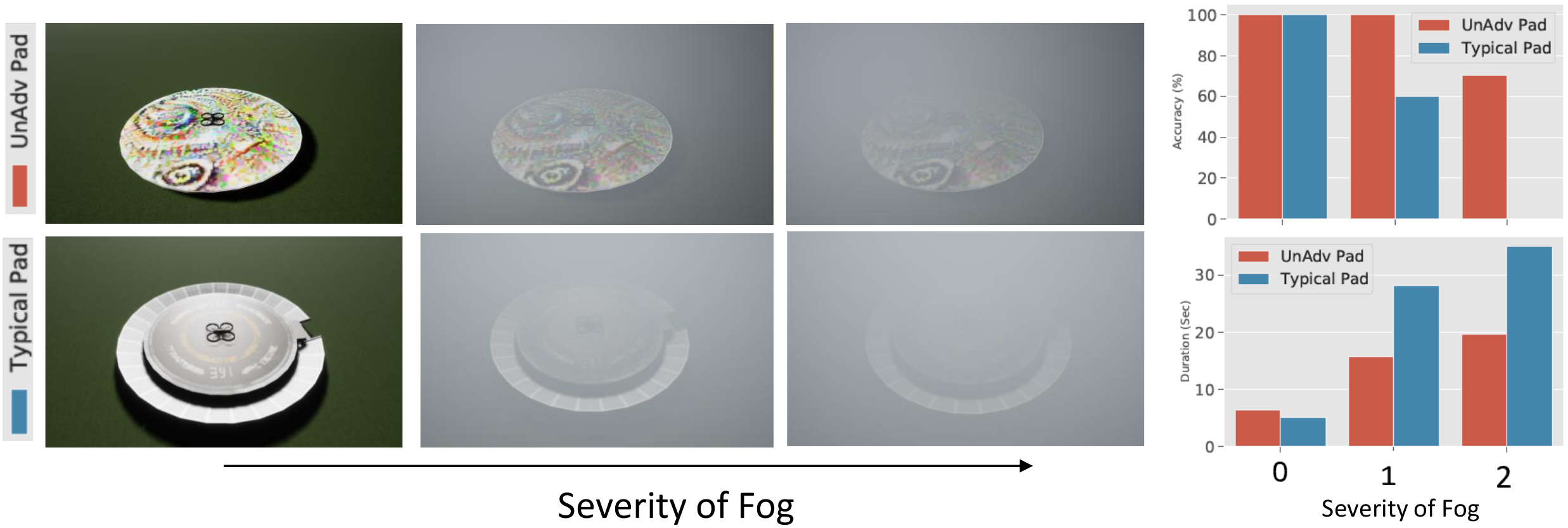}
    \label{fig:droneeval}
    \caption{Drone landing task. On the left we show the unadversarial versus
    standard landing pads. On the right we show the results for the task when
    both the standard and unadversarial landing pads are used. We find that the
    drone consistently takes less time to land, and has a higher chance of landing
    correctly, when detecting the unadversarial landing pad.}
    \label{fig:drone}
\end{figure*}

We then assess whether
unadversarial examples can help outside of the classification setting. Again
using AirSim, we set up a drone landing task with a perception module that
receives as input an axis-aligned aerial image of a landing pad, and is tasked
with outputing an estimate of the camera's $(x,y)$-position relative to the pad.
While this task is quite basic, we are particularly interested in studying
performance in the presence of heavy (simulated) weather-based corruptions.
The drone is equipped with a pretrained regression model that localizes
the landing pad (described in detail in Appendix~\ref{app:airsim}). We optimize
an unadversarial texture for the surface of the landing pad to best help the
drone's regression model in localization. Figure~\ref{fig:drone} shows
an example of the landing pad localization task, along with the performance of
the unadversarial landing pad compared to the standard pad. 
The drone landing on the unadversarial pad consistently lands both more
reliably. 

\subsection{Physical World Unadversarial Examples}
\begin{figure}
    \begin{subtable}{0.33\linewidth}
        {\begin{tabular}{@{}lcccc@{}}
        \toprule
         Class             & No Patch             & Patch  \\ \midrule 
        ``racer''          & 22\%                 & 83\% \\
        ``eggnog''         & 22\%                 & 44\%  \\
        ``coffee pot''     & 39\%                 & 56\%   \\
        ``warplane''       & 67\%                 & 83\%            \\ \bottomrule
        \end{tabular}}
        \caption{Accuracy of pre-trained ResNet-18 on photographs of real world objects with and without patches.}
        \label{tab:physical_world}
    \end{subtable}
    \hspace{1em}
    \begin{subfigure}{0.6\linewidth}
        \centering
        \includegraphics[width=\linewidth]{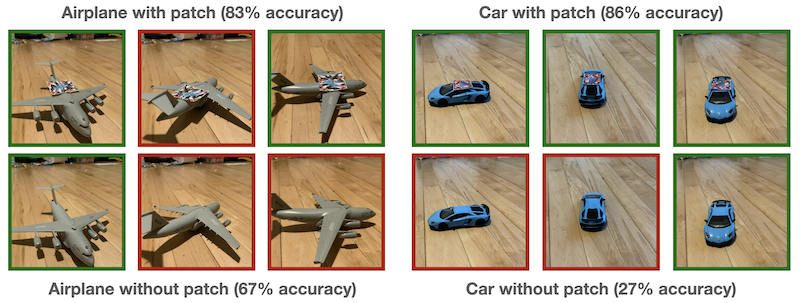}
        \caption{Examples photos of the ``warplane'' and ``racer'' physical
        objects taken with (top) and without (bottom) an unadversarial
        patch.}
        \label{fig:physical_world}
    \end{subfigure}
    \caption{Physical-world experiments. We take pictures of objects at diverse orientations while
    varying the presence of a patch on the object. Note that we don't do any
    additional data augmentation on the patches, which are the same used in
    our previous ImageNet benchmark experiment.}
\end{figure}

Finally, we move out of
simulation and test whether the unadversarial patches that we generate can
survive naturally-arising distribution shift from effects such as real lighting,
camera artifacts, and printing imperfections. We use four household objects (a
toy racecar, miniature plane, coffeepot, and eggnog container), and print out
(on a standard InkJet printer) the adversarial patch corresponding to the label
of each object. We take pictures of the toy with and without the patch taped on
using an ordinary cellphone camera, and count the number of poses for which the
toy is correctly classified by a pre-trained ImageNet classifier.
Our results are in Table~\ref{tab:physical_world}, and examples of patches
are in Figure~\ref{fig:physical_world}. Classifying both patched and
unpatched images over a diverse set of poses, we find that the adversarial
patches consistently improve performance even at uncommon object
orientations.

%% file: sections/related.tex
Here, we first highlight (and differentiate from) previous works using reference
markers to improve recognition and localization. We will then discuss related
work in adversarial robustness.

\paragraph{Improving computer vision with fiducial markers.} In past
research, vision-based precision landing was initially attempted with
classical computer vision based tracking of reference designs
\cite{saripalli2003visually, lange2009vision, yang2014autonomous,
falanga2017vision}, or on arrangements of fiducial markers
\cite{liu2019onboard} (a fiducial marker is a fixed pattern or object that is
placed in a scene as a reference point for location or measure). 
Several types of visual fiducial markers were also
proposed for robust tracking and pose estimation applications through
classical vision \citep{fiala2005artag, garrido2014automatic,
olson2011apriltag, romero2019fractal}. While fiducial markers such as
AprilTags \cite{olson2011apriltag} were commonly used in robotics, they
suffer from limitations such as orientation uncertainty, accuracy falloff
depending on viewing angles, and short detection ranges
\citep{abbas2019analysis, wang2016apriltag}, which motivated research into
using convolutional neural networks for landing pad detection and pose
estimation \citep{nguyen2018lightdenseyolo, truong2019deep}. Our work attempts
to unify these perspectives by leveraging the expressivity of neural
network-based systems to design robust unadversarial examples. Another key
difference between fiducial markers and unadversarial objects is that the former
require vision systems to be aware of their presence; in contrast, the latter
are designed on top of pre-trained systems. This means that we do not
require any further non-standard model training, and that we do not
depend as heavily on the unadversarial example being in the field of view.

\paragraph{Adversarial robustness.} Our design of unadversarial examples is
motivated by the success of adversarial examples, i.e., minute perturbations to
the inputs of machine learning models that can induce nearly arbitrary behavior.
Adversarial examples were shown to be effective in
seriously hampering machine vision related tasks such as classification
\cite{szegedy2014intriguing, evtimov2018robust, athalye2018synthesizing}, object detection/segmentation
\citep{eykholt2018physical, arnab2018robustness, lee2019physical,
xie2017adversarial} and visual question-answering (VQA) \cite{xu2018fooling}.
Furthermore, prior work has shown that adversarial examples can be constructed
even without direct access to the vision system being manipulated
\citep{chen2017zoo,papernot2016transferability,papernot2017practical,ilyas2018blackbox,ilyas2018prior}
Synthesized physical adversarial examples were shown to be effective in
fooling person detection \cite{thys2019fooling}, sign detection for
autonomous driving \citep{sitawarin2018darts, boloor2019simple}. Robotic
platforms such as manipulators were also shown to be sensitive to vision
based adversarial examples \cite{melis2017deep}, and to specifically designed
adversarial objects \cite{wang2019adversarial}. Additionally, a recent line of
work \cite{elsayed2018adversarialreprogramming,neekhara2019adversarial} shows
that one can ``reprogram'' neural networks using adversarial examples, e.g.,
one can construct an adversarial patch that when applied, causes a CIFAR-10
classifier to operate as an MNIST classifier.

%% file: sections/conclusions.tex
In this work, we demonstrated that it is possible to design object
alternations that boost the corresponding classifiers' performance, even
under strong and {\emph unforeseen} distribution shift. Indeed, such
resulting unadversarial objects are robust to a broad range of data shifts
and corruptions, even when these were never seen in training. We view our
results as a promising route towards increasing out-of-distribution
robustness of computer vision models.

\paragraph{Domains beyond image classification.}
The fact that unadversarial examples and adversarial examples share the same
underlying generation technique is evidence that unadversarial examples could
apply to any system that is vulnerable to adversarial examples. This indicates that
unadversarial examples could apply to a wide variety of
systems~\citep{carlini2016hidden,serban2020adversarial,suciu2019exploring,
kos2017adversarial,ebrahimi2018hotflip}.

\paragraph{Extensions} 
The method we present in this paper does not modify the underlying training
procedure at all, or modify the network used in conjunction with the object.
This presents an advantage in one sense, since we can use any pre-trained vision
model, but
it is possible that adding additional train
time augmentations could make the unadversarial (object, model) pair more robust
at completing the target
task. 
For example, we could use
data augmentation or train objects on the actual distribution shift we want to 
be robust to if it is specifiable.

\paragraph{Limitations}
One limitation of our method is that it requires differentiability with
respect to properties of the input object of interest, or a close proxy (e.g.
a differentiable simulator) that can mimick the object and the environment in
which the object operates.

%% file: appendix/appendix.tex
\appendix
\onecolumn
\section{Pseudocode for Unadversarial Example Generation}
\label{app:algorithms}
\input{appendix/algorithms}

\clearpage
\section{3D Simulation Details}
\label{app:airsim}
\input{appendix/airsim}

\clearpage
\section{Experimental Setup}
\label{app:exp_setup}
\input{appendix/exp_details}

\clearpage
\section{Omitted Results}
\label{app:detailed_results}
\input{sections/detailed_results}

%% file: appendix/algorithms.tex
\begin{algorithm}[H]
    \SetAlgoLined
    \KwIn{Pre-trained classifier with parameters $w$, loss function $\ell_w(x, y)$, dataset $\mathcal{D}$}
    \KwIn{Image size $m$, patch size $n$, target class $C_{targ}$, patch learning rate $\eta$}
    \KwResult{An unadversarial patch for the class $C_{targ}$}
    Randomly initialize a patch $\theta \in \mathbb{R}^{n \times n \times 3}$\;
    \For{$K$ iterations}{
    Sample batch of image-label pairs $(x, y) \sim \mathcal{D}$\;
    \If{$y = C_{targ}$}{
    $\theta_{padded} \gets $ Zero-pad $\theta$ to size $m \times m$\;
    {\em mask} $\gets \text{int}(\theta_{padded} > 0)$ \tcp*{0/1 mask signalling patch location}
    $T \gets $ RandomAffineTransform() \tcp*{random rotation, scaling, and translation}
    $x_{unadv} \gets x \cdot (1 - T(mask)) + T(\theta_{padded}) \cdot T(mask)$
    \tcp*{apply patch using mask}
    $\theta \gets \theta - \eta \cdot \text{sign}\left(\nabla_{\theta}
    \ell_w(x_{unadv}, y)\right)$ \tcp*{gradient descent step on $\theta$}
        }
    }
    \Return{$\theta$}
    \caption{Unadversarial patch generation}
    \label{alg:patches}
\end{algorithm}

\begin{algorithm}[H]
    \SetAlgoLined
    \KwIn{Pre-trained classifier with parameters $w$, loss function $\ell_w(x,
    y)$, set of background images $\mathcal{D}_b$}
    \KwIn{Texture size $n$, target 3D mesh $M_{targ}$, texture learning rate $\eta$}
    \KwResult{An unadversarial texture for the mesh $M_{targ}$}
    Randomly initialize a texture $\theta \in \mathbb{R}^{n \times n \times 3}$\;
    Init a texture $t_{uv} \in \mathbb{R}^{n \times n \times 3}$ with
        $t_{uv}[i,j,1] = i$, $t_{uv}[i,j,2] = j$, $t_{uv}[i,j,3] = 0$ \tcp*{$t_{uv}$ 
        is a {\em UV map}}
    \For{$K$ iterations}{
    Sample background $x_{bg} \sim \mathcal{D}$\;
    Sample a random 3D configuration (position and orientation) $C_{3D}$\;
    $x_{rend} \gets$ render $M_{targ}$ 
        in configuration $C_{3D}$ with texture $\theta$ and background $x_{bg}$\;
    $x_{uv} \gets$ render $M_{targ}$ in configuration $C_{3D}$ with texture
        $t_{uv}$ and clear background\;
    $x_{drend} \gets$ linear (differentiable) approximation to $x_{rend}$, i.e.,
    \[
        x_{drend}[i,j] = \begin{cases}
            x_{bg}[i,j] &\text{if $x_{uv}[i,j]$ is blank} \\
            \theta[x_{uv}[i,j]] &\text{if $x_{uv}[i,j]$ is not blank}
        \end{cases}
    \]

    $x_{unadv} \gets x_{drend} - \text{detach}(x_{drend}) + x_{rend}$ 
    \tcp*{so $x_{unadv} = x_{rend}$ but $\nabla_\theta x_{unadv} =
    \nabla_\theta x_{drend}$}
    $\theta \gets \theta - \eta \cdot \text{sign}\left(\nabla_{\theta}
    \ell_w(x_{unadv}, y)\right)$ \tcp*{gradient descent step on $\theta$}
    }
    \Return{$\theta$}
    \caption{Unadversarial texture generation}
    \label{alg:textures}
\end{algorithm}

%% file: appendix/airsim.tex
\subsection{Overview of AirSim} We conduct our simulation experiments using
the high fidelity simulator, Microsoft AirSim. AirSim acts as a plugin to
Unreal Engine, which is a AAA videogame engine providing access to high
fidelity graphics features such as high resolution textures, realistic
lighting, soft shadows etc. making it a good choice for rendering for
computer vision applications. AirSim internally provides physics models for a
quadrotor vehicle, which we leverage for performing autonomous drone landing.
As a plugin, AirSim can be paired with any Unreal Engine environmnent to
simulate autonomous vehicles that can be programmed with an API both in terms
of planning/control as well as obtaining camera images. AirSim also allows
for controlling environmental features such as time of day, dynamically
adding/removing objects, changing object textures and so on.

\subsection{3D Boosters Classification Experiment} 
\paragraph{Format of 3D models}
To evaluate the performance of pretrained ImageNet classifiers at detecting
3D unadversarial/boosted objects (e.g. the jet shown in the main paper) among
realistic settings, we set up an experiment using AirSim for image
classification of common classes (warplane, car, truck, ship, etc). We pick
the class of `warplane' as our object class of interest download publicly
available 3D meshes for this class from \url{www.sketchfab.com}. Using the
open source 3D modeling software Mitsuba, we modify the object texture to
match the boosted texture for the corresponding class, and then export these
meshes into the GLTF format for ingestion into Unreal Engine/AirSim. This
allows us to import the boosted objects into the AirSim framework, and spawn
them as objects in any of the environments being created.

\begin{figure*}[!htbp] \begin{subfigure}[t]{0.22\textwidth} 
    \centering
    \includegraphics[width=\textwidth]{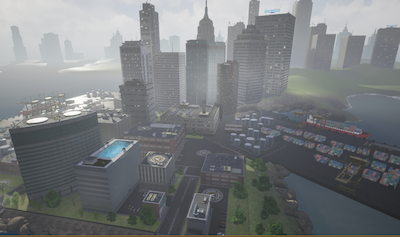}
    \caption{City environment in AirSim used for detection experiment}
    \label{fig:city} 
\end{subfigure} 
\quad 
\begin{subfigure}[t]{0.22\textwidth}
    \centering
    \includegraphics[width=\textwidth]{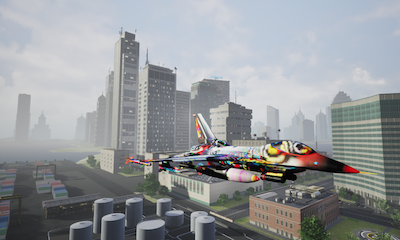}
    \caption{Boosted `jet' model in the City environment.} 
    \label{fig:jet_city}
\end{subfigure} 
\quad 
\begin{subfigure}[t]{0.23\textwidth} 
    \centering
    \includegraphics[width=\textwidth]{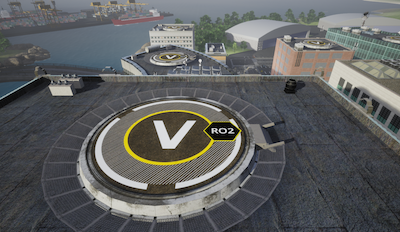}
    \caption{Sample landing pads atop buildings in the City environment.}
    \label{fig:pad_city} 
\end{subfigure} 
\quad
\begin{subfigure}[t]{0.23\textwidth} 
    \centering
    \includegraphics[width=\textwidth]{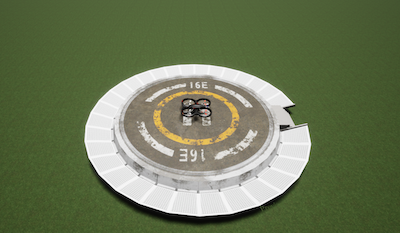}
    \caption{Drone in test environment used for the landing experiments.}
    \label{fig:pad_grass} 
\end{subfigure} 
\end{figure*}

\paragraph{Environment screenshots and description} Within AirSim, in the
interest of generating realistic imagery, we simulate a city environment
(\autoref{fig:city}). For this experiment, we use the ComputerVision mode of
AirSim, which does not simulate a vehicle but rather, gives the user control
of a free moving camera, allowing us to generate data at ease from various
locations and varying camera and world parameters.

\paragraph{Sampling and evaluation} Once the 3D objects (unadversarial or
normal) are present in AirSim's simulated world, the next step is to evaluate
the classification of these objects from different camera angles, weather
conditions etc. Given the location of a candidate object (which we randomize
and average over five locations), we sample a grid ($10\times10\times10$) of
camera positions in 3D around the object. For each of these positions, we
move AirSim's main camera and orient it towards the object, resulting in
images from various viewpoints. At runtime, each of these images are
immediately processed by a pretrained ResNet-18 ImageNet classifier, which
reports the top 5 class predictions. We average the accuracies across the
five different locations in the scene and the $1000$ grid points around the
object at each location.

Along with this variation in camera angles and thereby, object pose in the
frame; we also evaluate the performance of of the various 3D objects given
environmental perturbations. We achieve this through the AirSim's weather
conditions feature, using which we simulate weather conditions such as dust
and fog dynamically with varying levels of severity of these conditions. We will
open-source binaries for the AirSim code and environments that we use which will
allow people to replicate our results, and investigate more scenarios of
interest.

\subsection{Drone Landing Experiment}

In this experiment, we evaluate how unadversarial/boosted objects can help
robustify perception-action loops that are driven by vision-based pose
estimation. Perception-action loops are at the heart of many robotics tasks,
and accurate perception is imperative for safe, efficient navigation of
robots. We choose the scenario of autonomous drone landing as our experiment,
and simulate it within AirSim.

For this experiment, we create assets of landing pads that are similar to
helipads on top of buildings in the city environment(\autoref{fig:pad_city}).
We also use a test environment with a single landing pad located on a patch
of grass. An example of such a landing pad can be seen in
\autoref{fig:pad_grass}. We use AirSim to simulate a quadrotor drone in these
worlds, which can be programmatically controlled using a Python API. AirSim
allows us to equip a downward facing, gimballed camera on this drone in order
to obtain RGB images, which are then processed by our landing pad pose
estimation (regression) model. Given an RGB image, the regression model
outputs a 6 degree of freedom pose for the landing pad. We use/optimize only
the first two enteries of this output corresponding to the relative x and y
location of the landing pad w.r.t the drone.

We formulate the drone landing experiment as a visual servoing task: a
perception action loop that involves estimating the relative location of the
pad from the image frame captured by the downward facing camera of the drone,
and sending an appropriate velocity command in order to align the camera
center with that of the pad. We achieve these through the following steps:

\paragraph{Data Collection.} We use AirSim's inbuilt data collection API for
this step. Given the location of the pad in the world, we sample various
feasible locations for the drone in an imaginary cone whose vertex aligns
with the center of the landing pad. We then spawn the drone in these randomly
sampled positions, and obtain the RGB and segmentation views of the pad as
generated by AirSim, along with the relative ground truth position of the
landing pad w.r.t the drone, and repeat this process to create a dataset. The
collected dataset contains 20000 images and is split 80-20 between train and
evaluation sets.

\paragraph{Landing pad pose estimator.} We train a model that maps top view
images of a scene with a landing pad, to the relative 2D location of the
landing pad w.r.t the drone in the camera frame. We use a ResNet-18
pretrained on ImageNet as the backbone for the pose regressor, and we replace
the last classification layer with a regression layer outputting the $(x,y)$
relative location of the pad w.r.t drone. The model is trained end-to-end by
minimizing the mean squared error (MSE) loss between the predicted location
and the ground truth location. The ground truth is collected along with the
images using the AirSim City simulation environment as describe before.

We train the model for 10 epochs using SGD with a fixed learning rate of
0.001, a batch size of 512, a weight decay of 1e-4, and with MSE as the
objective function. The model converges fairly quickly (within the first few
epochs). 

\paragraph{Drone Landing.} To use the pose estimator's predictions and send
appropriate actions, we utilize the Multirotor API of AirSim. This allows us
to control the drone by setting the desired velocity commands along all the
axes (translational/rotational). Given the position of the landing pad in the
scene relative to that of the drone( as output by the pose regressor) we
execute the landing operation by sending appropriate velocity commands to the
drone.

To generate the right velocity commands, given the relative position of the
landing pad, we use a standard PID controller that computes corrective
velocity values until the position of the drone matches that of the landing
pad. For a pose output by the regressor treated as the setpoint $P_{set}$,
and current drone pose $P_{curr}$ and at any point at time $t$, the
appropriate velocity command $v(t)$ can simply be computed as follows:
\begin{align} e(t) &= P_{set} - P_{curr} \nonumber \\\nonumber v(t) &= K_p
e(t) + K_d \frac{d}{dt}e(t) + K_i * \int_{0}^{t} e(t)dt \end{align} where
$K_p, K_d, and K_i$ are the hyperparameters of the PID controller and are
manually tuned. We find that $K_p = K_d = 5$ and $K_i = 0$ to be reasonable
for our task.

For realistic perturbations to the scene, similar to the 3D boosters
classification experiment, we continue making use of the weather API to
generate weather conditions in AirSim. This results in a variation of factors
such as amount of dust or fog in the scene, allowing us to evaluate the
performance of landing under various realistic conditions.

%% file: appendix/exp_details.tex
\subsection{Pretrained vision models we evaluate}
\label{app:pretrained-models} Here we present details of the different vision
models we use in our paper. For more details on all of these, please check
the README of our code at \codelink.

\paragraph{Corruption benchmark experiments:} We use pretrained ResNet-18 and
ResNet-50 (both standard and $\ell_2$-robust with $\varepsilon = 3$)
architectures from \cite{salman2020adversarially}:
\url{https://github.com/microsoft/robust-models-transfer}. \paragraph{3D
object classification in AirSim:} We use an ImageNet pretrained ResNet-18
architecture from the PyTorch's Torchvision\footnote{These models can be
found here \url{https://pytorch.org/docs/stable/torchvision/models.html}} to
classify all the boosted and non-boosted versions of the jets, cars, ships
etc in AirSim.

\paragraph{Drone landing experiment in AirSim:} We finetune an ImageNet
pretrained ResNet-18 model on the regression task of drone landing. The last
layer of the pretrained model is replaced with a 2D linear layer estimating
the relative pad location w.r.t the drone. We collect a 20k sample dataset
for training the pad pose estimation in AirSim with an $80-20$ train-val
spilt. We use a learning rate of $0.001$, a batch size of $512$, a weight
decay of $1e-4$. We train for $10$ epochs.

\paragraph{Physical world unadversarial examples experiment:} Similar to the
3D object classification experiment in AirSim, we use an ImageNet pretrained
ResNet-18 architecture from Torchvision.

\subsection{Unadversarial patch/texture training details} \paragraph{Patches
training details} We fix the training procedure for all of the 2D patches we
optimize in our paper. We train all the patches starting from random
initialization with batch size of 512, momentum of $0.9$, and weight decay of
$1e-4$. We train all the patches for 30 epochs (which is more than enough as
we observe that for both ImageNet and CIFAR-10, the patch converges within
the first 10 epochs) with a learning rate of $0.1$ {We sweep over three
learning rates $\in \{0.1, 0.01, 0.001\}$ but we find that all of these
obtain very similar results. So we stick with a learning rate of $0.1$ for
all of our experiments.}.

For the classification tasks (i.e., everything but drone landing) we
use the standard cross-entropy loss. For the drone landing task (landing pad
pose estimation), we use the standard mean squared error loss.

\paragraph{Texture training details} We now outline the process for
constructing adversarial textures. We implemented a custom PyTorch module
with a distinct forward and backward pass; on the forward pass (i.e., during
evaluation), the module takes as input an ImageNet image, and a 200px by
200px texture; using the Python bindings for Mitsuba~\citep{mitsuba} 3D
renderer, the module returns a rendering of the desired 3D object, overlaid
onto the given ImageNet image. On the backwards pass (i.e., when computing
gradients), we use the 3D model's UV map\footnote{Mitsuba provides direct
access to the UV map through the \texttt{aov} integrator; see the code
release for more details.}---a linear transformation from $(x, y)$ locations
on the texture to $(x, y)$ locations in the rendered image---to approximate
gradients through the rendering process. This is the same procedure used
by~\citep{athalye2018synthesizing} for constructing physical adversarial
examples. Note that this is a simple approximation that only accounts for the
location of pixels in the rendered image (i.e., ignores the effects of
lighting, warping, etc.). However,

\subsection{Details of the physical world experiment} 
\begin{figure}
    \centering
    \includegraphics[width=\textwidth]{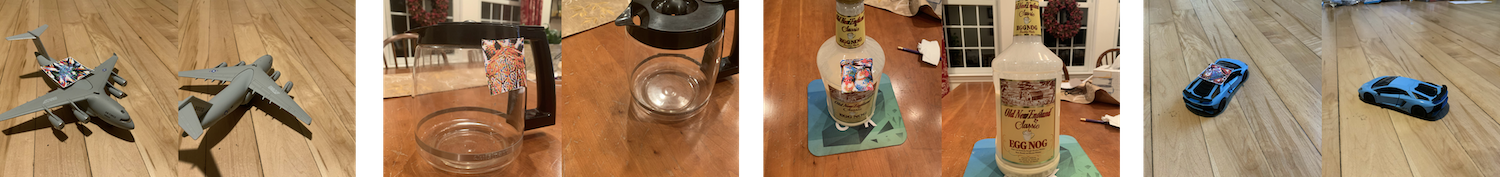}
    \caption{Photographs in different poses of the four physical objects we
    experimented on, with and without an unadversarial patch.}
    \label{fig:more_physical_examples}
\end{figure}

To conduct the
physical-world experiments, we used a toy
racecar\footnote{https://www.amazon.com/gp/product/B07T5X69TZ/}, a toy 
warplane\footnote{\url{https://www.amazon.com/CORPER-TOYS-Pull-Back-Aircraft-Birthday/dp/B07DB3839X/}}
(both from \url{amazon.com}) as well as two household objects: a coffeepot and
eggnog container. We then printed the unadversarial patches
corresponding to classes ``racer,'' ``warplane,'' ``coffeepot,'' and ``eggnog'' on an HP DeskJet
2700 InkJet printer, at 250\% scale. We adhere the patches to the top of
their respective objects with clear tape (the results are shown in
Figure~\ref{fig:physical_world}). We choose 18 distinct poses (camera
positions), and for each pose took one picture of the object with the patch
attached, and one picture without (keeping the location of the patch constant
throughout the experiment). Example photographs are shown in Figure
\ref{fig:more_physical_examples}. We evaluated a pre-trained ResNet-18
classifier on the resulting images.

\subsection{Replicate our results} We desired simplicity and kept
reproducibility in our minds when conducting our experiments, so we use
standard hyperparameters and minimize the number of tricks needed to
replicate our results. Our code is available at \codelink.

%% file: sections/detailed_results.tex
\begin{figure}[!htbp]
    \centering
    \includegraphics[width=.415\linewidth]{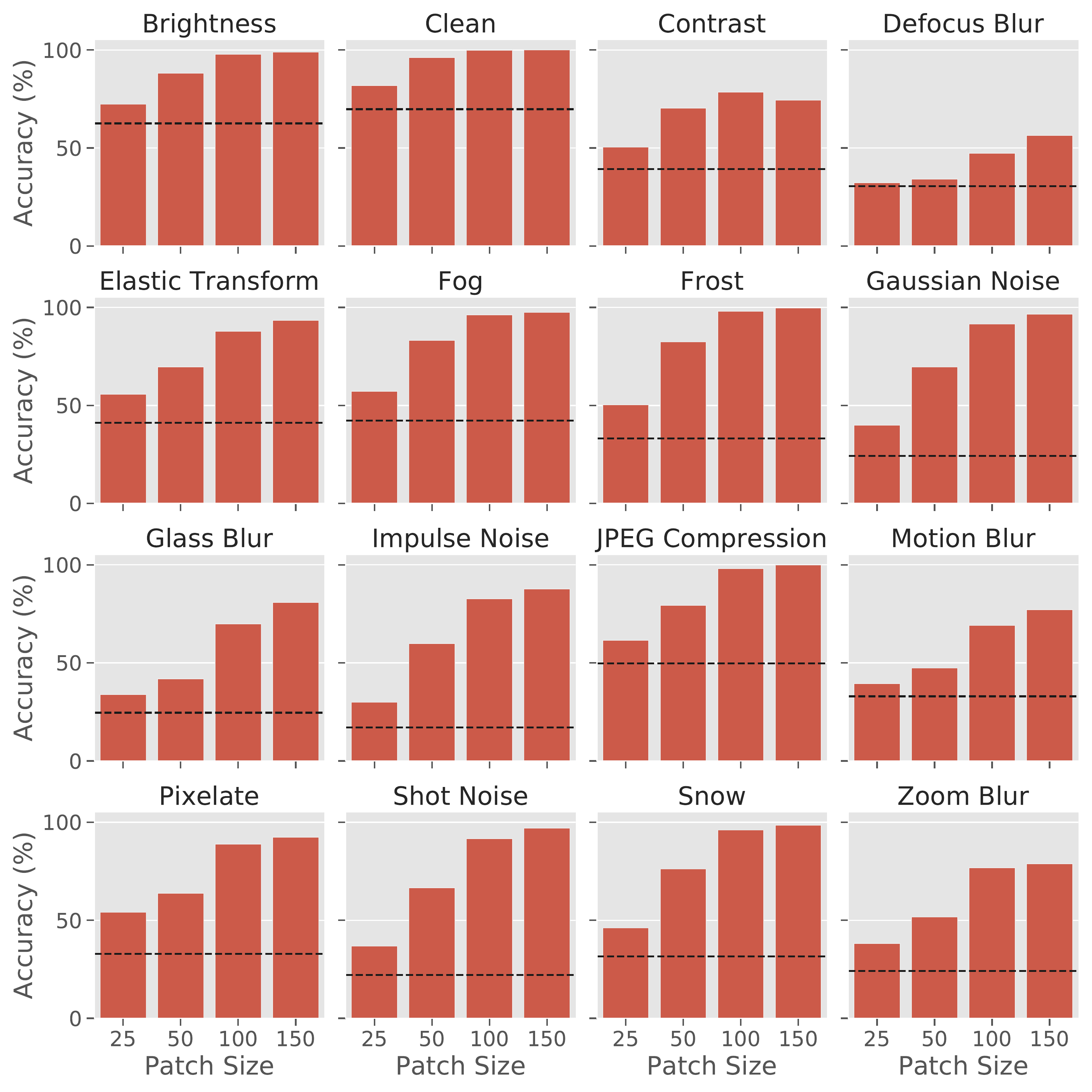}        
    \includegraphics[width=.49\linewidth]{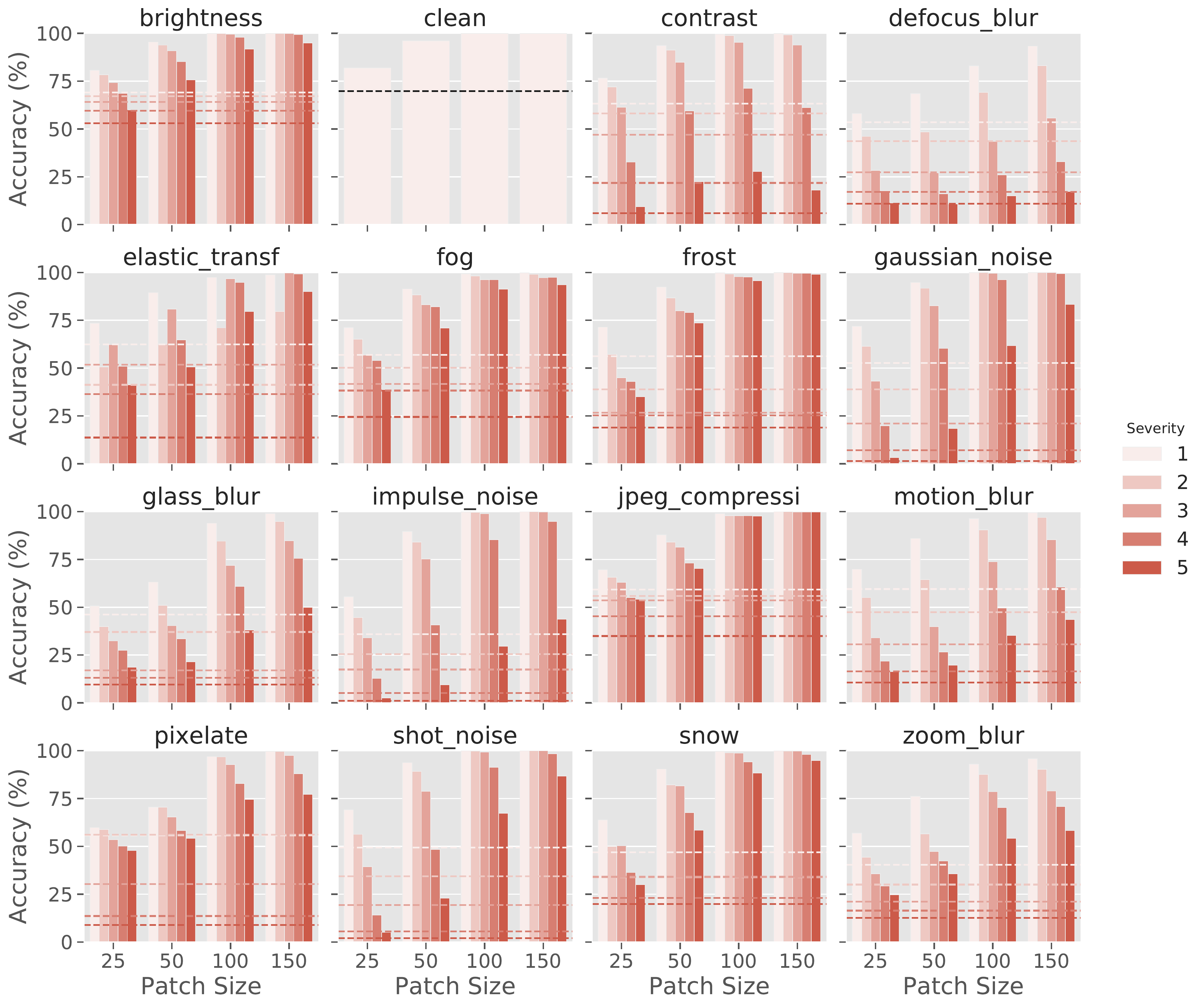}        
    \caption{Robustness of a trained 2D booster over pretrained ImageNet ResNet-18 model.} 
    \label{fig:imagenet-main-results-app}
\end{figure}
\begin{figure}[!htbp]
    \centering
    \includegraphics[width=.415\linewidth]{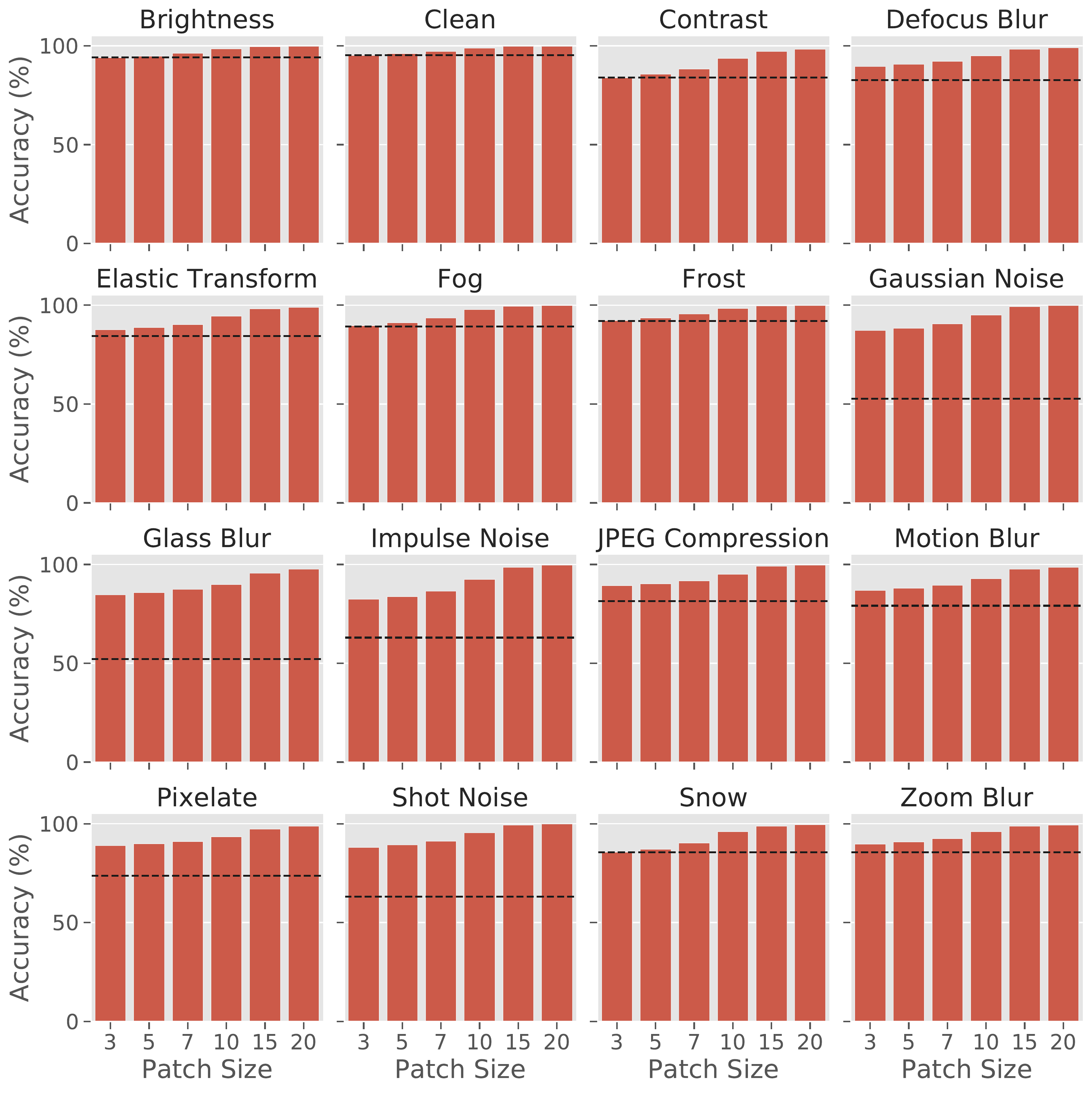}        
    \includegraphics[width=.49\linewidth]{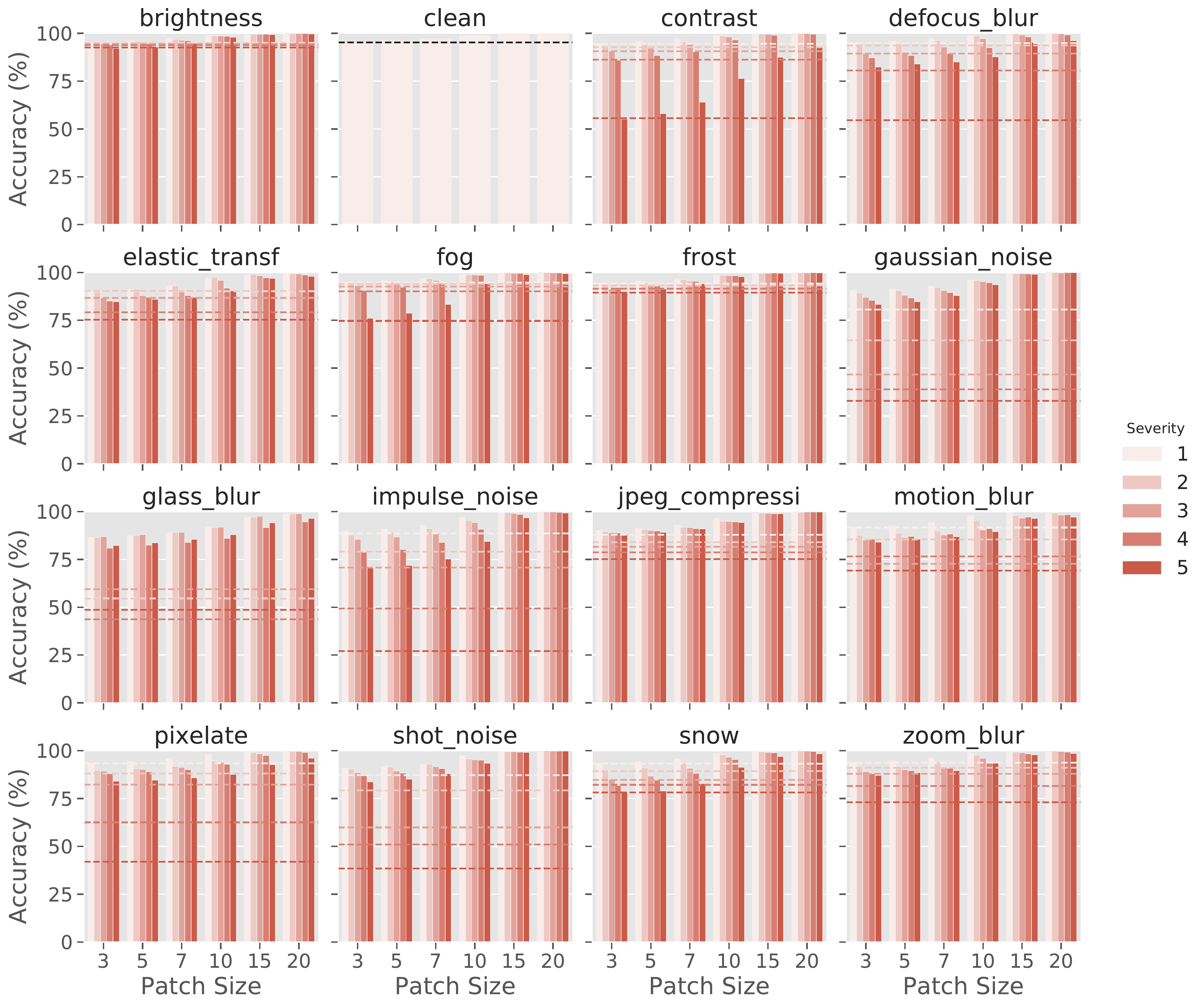}        
    \caption{Robustness of a trained 2D booster over pretrained CIFAR-10 ResNet-50 model.} 
    \label{fig:cifar10-main-results}
\end{figure}

In the below figure, we show a more detailed look of the main results of the benchmarking experiments in our paper, along with useful baselines. The single color plots (e.g. the left subplot in \autoref{fig:imagenet-main-results-app}) report the average performance over the 5 severities of ImageNet-C and CIFAR-10-C. The multicolor bar plots (e.g. the right subplot in \autoref{fig:imagenet-main-results-app}) report the detail performance per severity level. The horizontal dashed lines report the performance of the pretrained models on the original (non-patched) ImageNet-C and CIFAR-10-C datasets and serve as a baseline to compare with. For both ImageNet and CIFAR as shown in \autoref{fig:cifar10-main-results} and \autoref{fig:imagenet-main-results-app}, we are able to train unadversarial patches of various size that, once overlaid on the datasets, make the pretrained model consistently much more robust under all corruptions.

\subsection{Corruption benchmark main results: additional results to \autoref{fig:2d_results}}
Here we show the detailed main results for boosting ImageNet and CIFAR-10 with unadverasarial patches.

\clearpage
\subsection{Baselines}
\input{sections/baselines}

%% file: sections/baselines.tex
Below, we report a number of alternative ways to create patches for boosting the performance of object recognition.

\subsubsection{QR-Code}
\begin{figure}[!htbp]
    \centering
    \includegraphics[width=.8\linewidth]{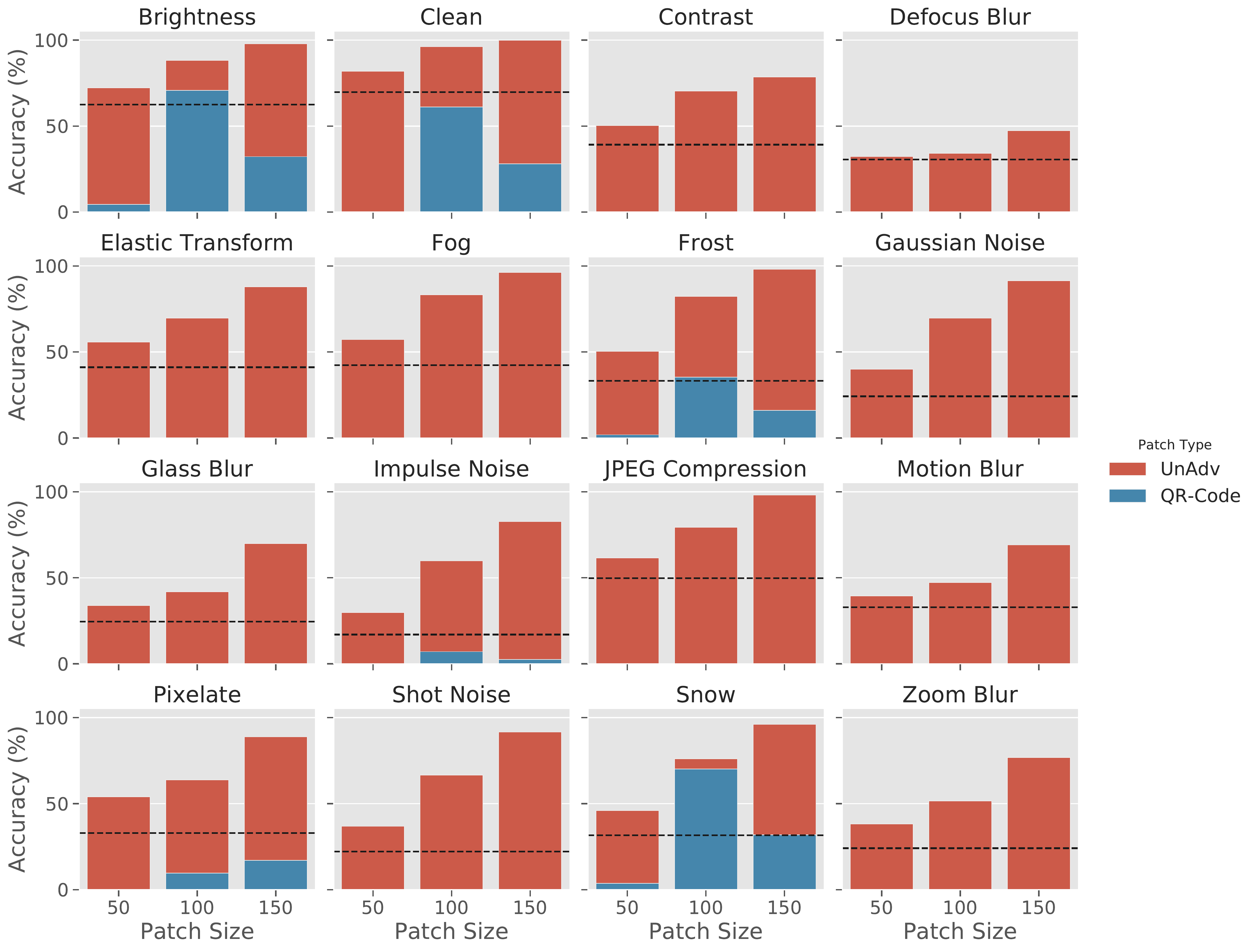}        
    \\
    \includegraphics[width=.8\linewidth]{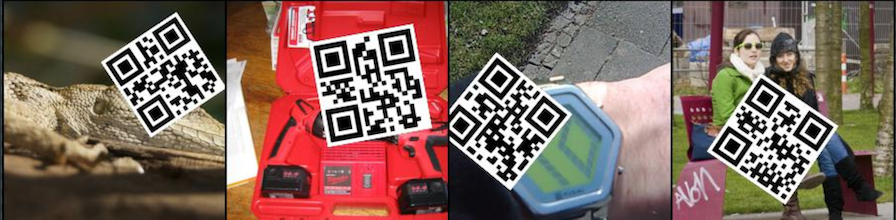}        
    \caption{QR-Code boosted ImageNet results under various corruptions.} 
    \label{fig:qrcode-results}
\end{figure}

We compare our unadversarial patches to the well-known QR-Code patches. We create a QR-Code for each class of the ImageNet dataset using Python's \texttt{qrcode} package(we avoid using CIFAR-10 since the images are too small for QR-Codes to be visible and detected at all). We overlay the QR-Codes over the ImageNet validation set according in accordance to what label each image has. We add the various ImageNet-C corruption on top of the resulting images, then we use python's \texttt{Pyzbar}\footnote{We experiment with \texttt{OpenCV for detecting the QR-Codes but find that \texttt{Pyzbar} leads to better performance}.} package to detect the QR-Codes. The results are shown in \autoref{fig:qrcode-results}. The performance of QR-Codes is not comparable to what we obtain with unadversarial patches (see \autoref{fig:2d_results}).

\clearpage
\subsubsection{Best training image per class as patch}
\begin{figure}[!htbp]
    \centering
    \includegraphics[width=.45\linewidth]{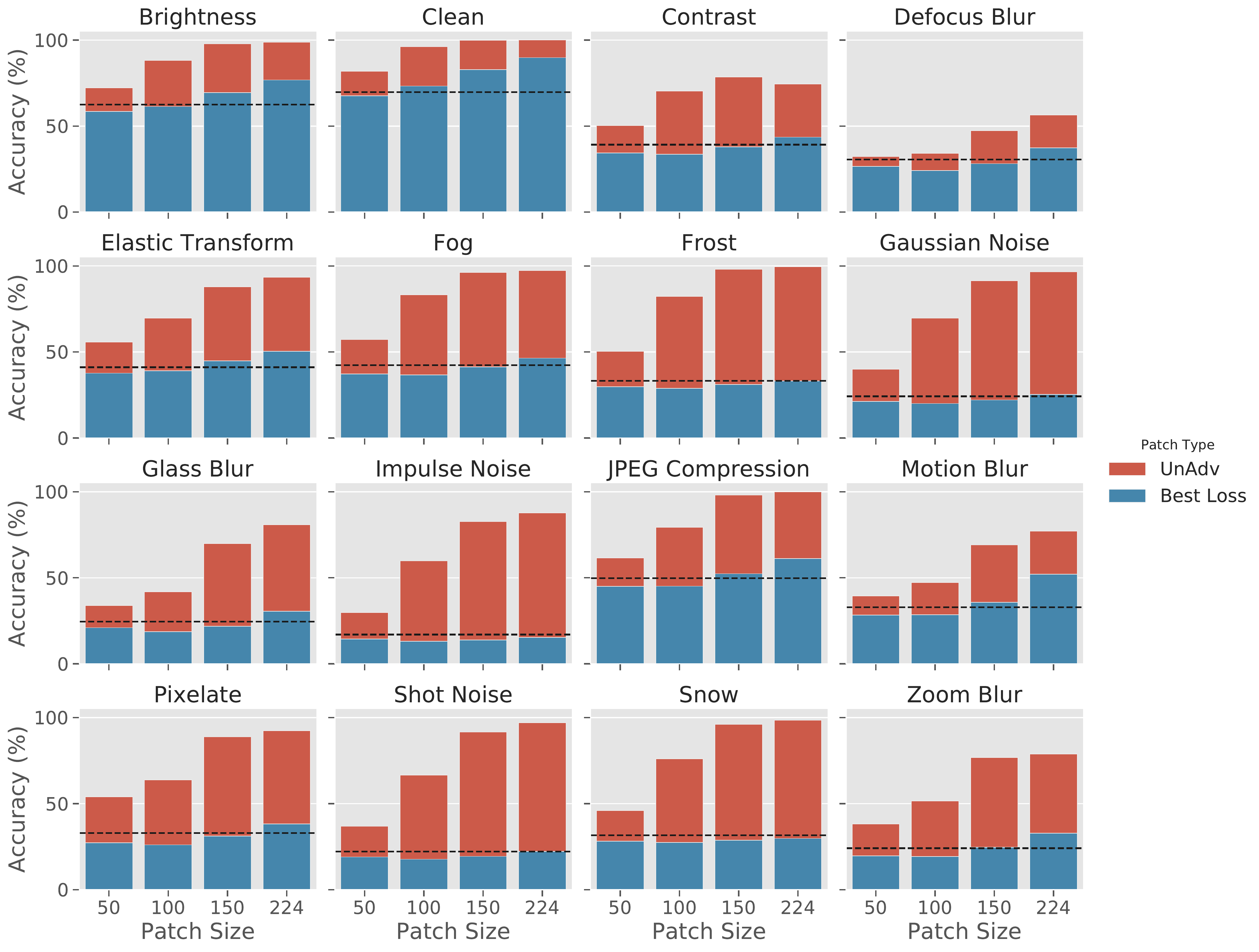}        
    \\
    \includegraphics[width=.4\linewidth]{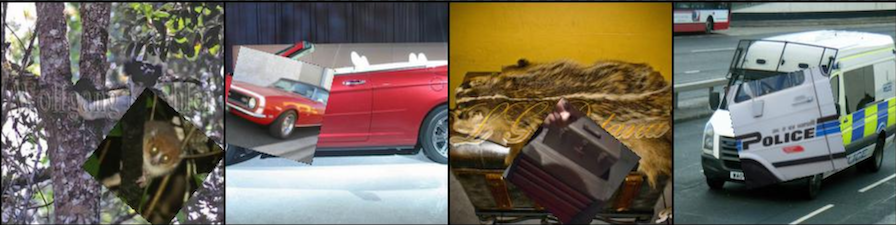}        
    \caption{Best training image with translation, rotation, and scaling for ImageNet.} 
    \label{fig:bestimages-imagenet}
\end{figure}

\begin{figure}[!htbp]
    \centering
    \includegraphics[width=.45\linewidth]{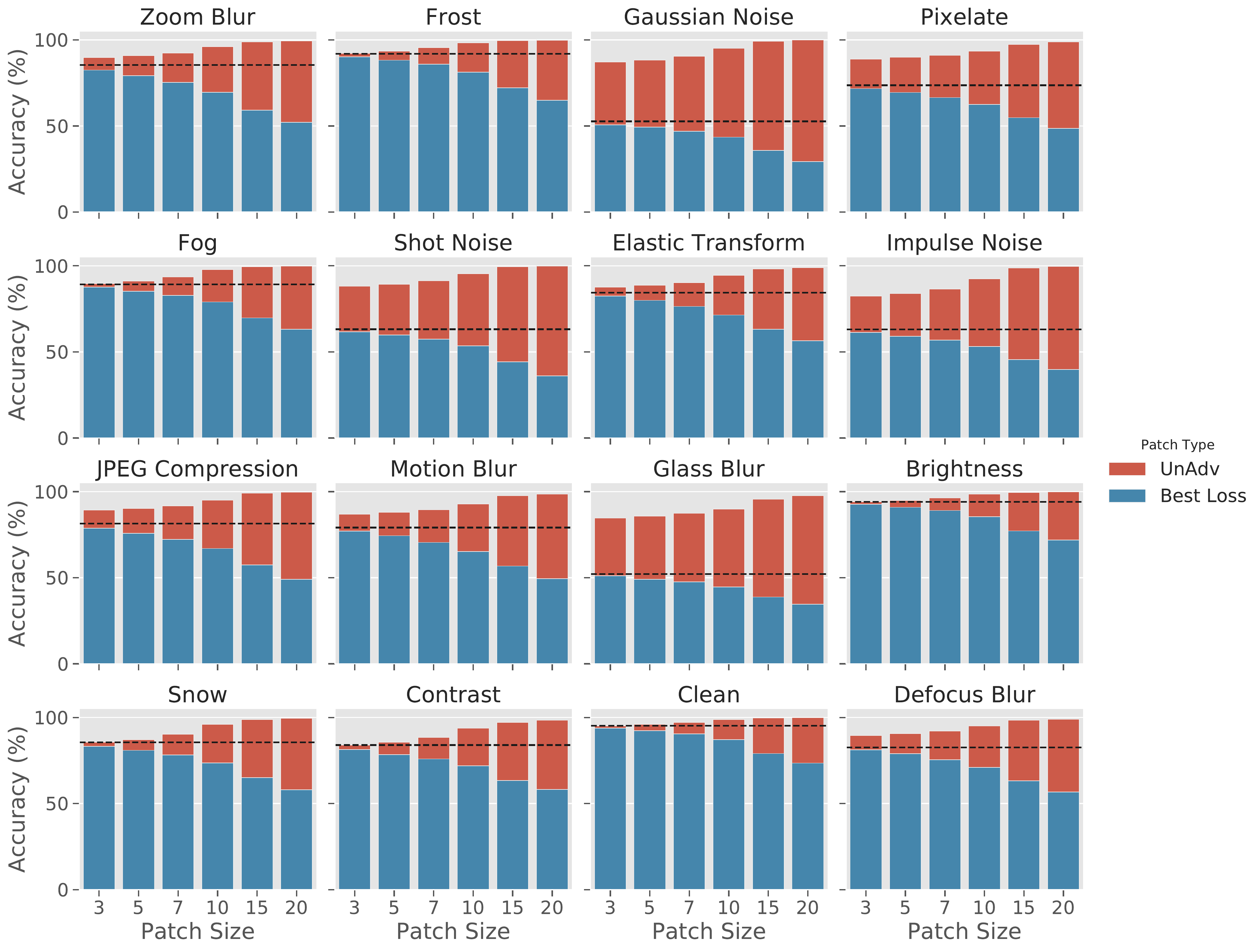} 
    \\       
    \includegraphics[width=.4\linewidth]{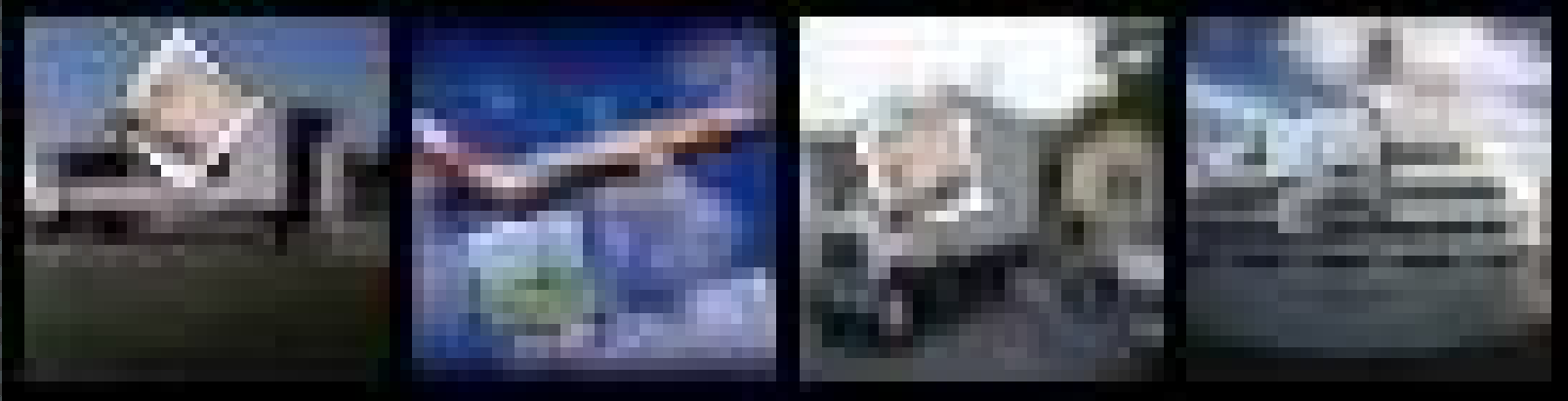}        
    \caption{Best training image with translation, rotation, and scaling for CIFAR-10.} 
    \label{fig:bestimages-cifar}
\end{figure}

Another natural baseline that we compare with is using the best images per class in the training set of the task of interest as patches for boosting the performance of pretrained models. For example, for ImageNet classification, we simply evaluate the loss of each training image using a pretrained ImageNet model (ResNet-18 in our case), and we the image with the lowest loss per class as the patch for that class. Now we overlay these found patches onto the ImageNet validation set with random scaling, rotation, and translation (as shown in \autoref{fig:bestimages-imagenet}), we add ImageNet-C corruptions, and we evaluate this new dataset using the same pretrained model we used to extract the patches. The results for ImageNet and CIFAR-10 are shown in \autoref{fig:bestimages-imagenet} and \autoref{fig:bestimages-cifar}, respectively.

\clearpage
\subsubsection{Best training image vs random training image as patch}
\begin{figure}[!htbp]
    \centering
    \includegraphics[width=.55\linewidth]{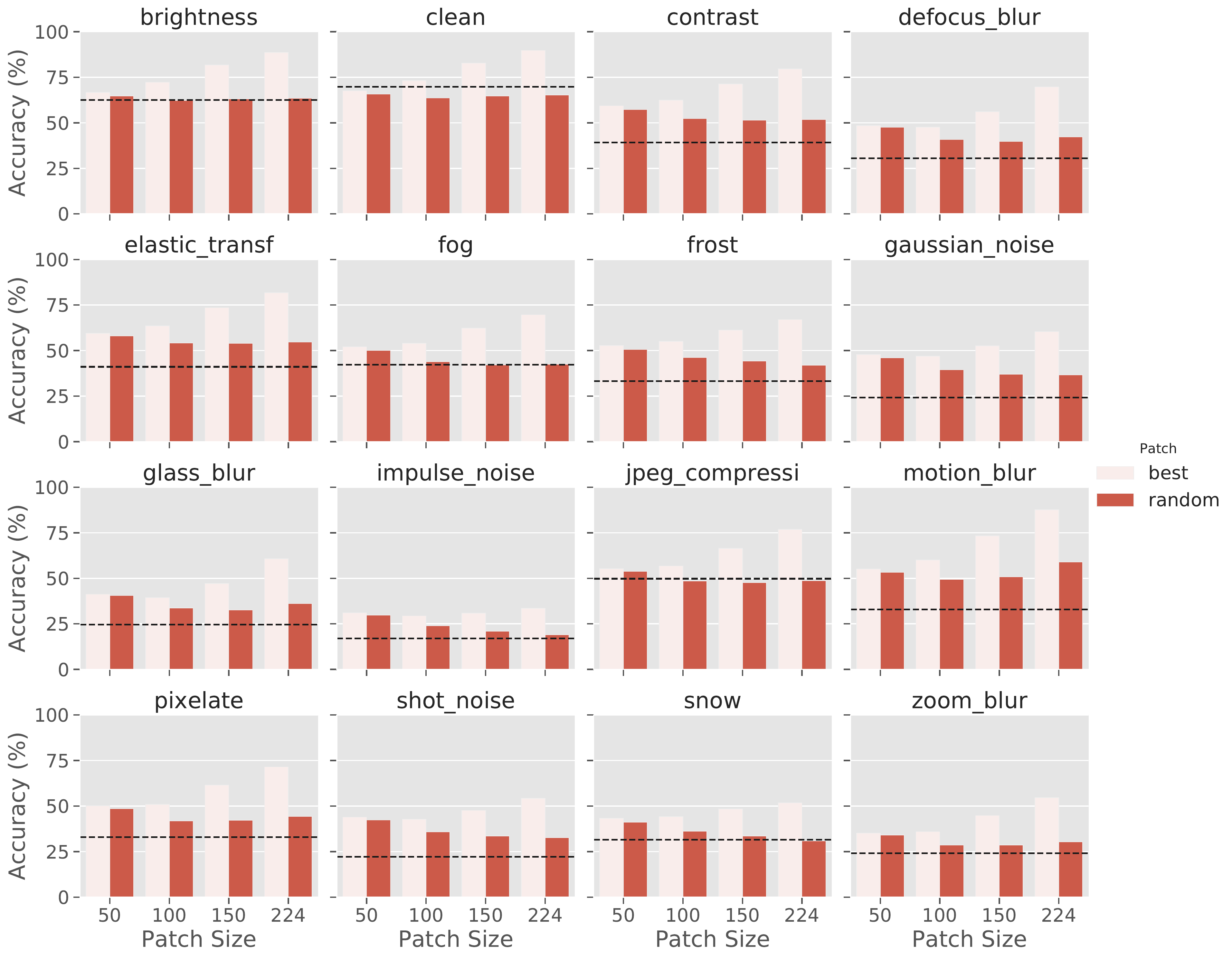}        
    \caption{Best training image vs random training image with translation, rotation, and scaling.}
    \label{fig:imagenet-rand-image-res} 
\end{figure}

\begin{figure}[!htbp]
    \centering
    \includegraphics[width=.55\linewidth]{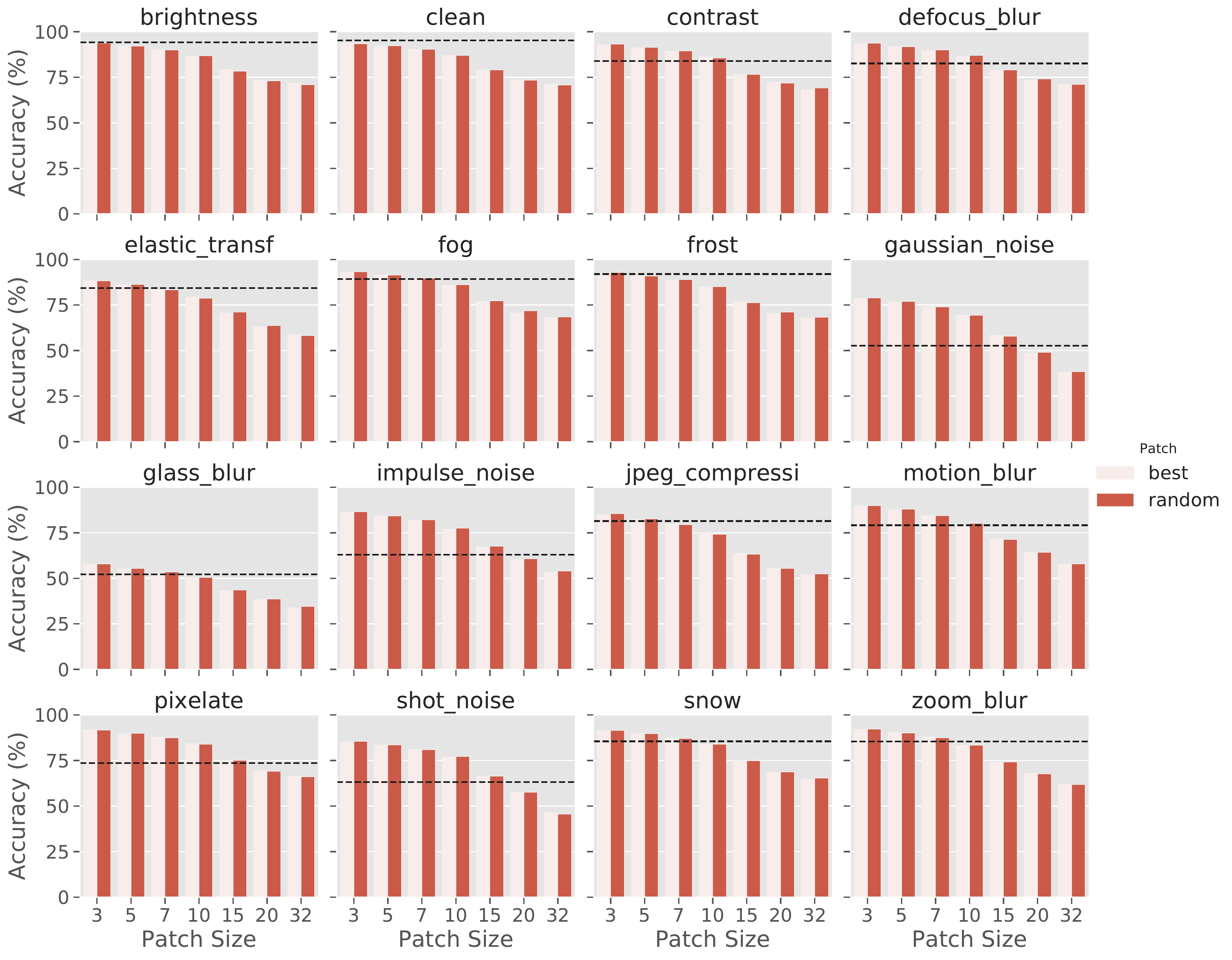}        
    \caption{Best training image vs random training image with translation, rotation, and scaling.} 
    \label{fig:cifar-rand-image-res}
\end{figure}

Here we investigate whether using a random image from the training set does any better than using the best-loss image as a patch. The results are shown in the below Figures. As one would expect, using a random image from the training set leads to strictly worse performance. This holds for both ImageNet and CIFAR-10 as shown in \autoref{fig:imagenet-rand-image-res} and \autoref{fig:cifar-rand-image-res}.

\clearpage
\subsubsection{Predefined fixed-pattern unadversarial patches}
\begin{figure}[!htbp]
    \centering
    \includegraphics[width=.6\linewidth]{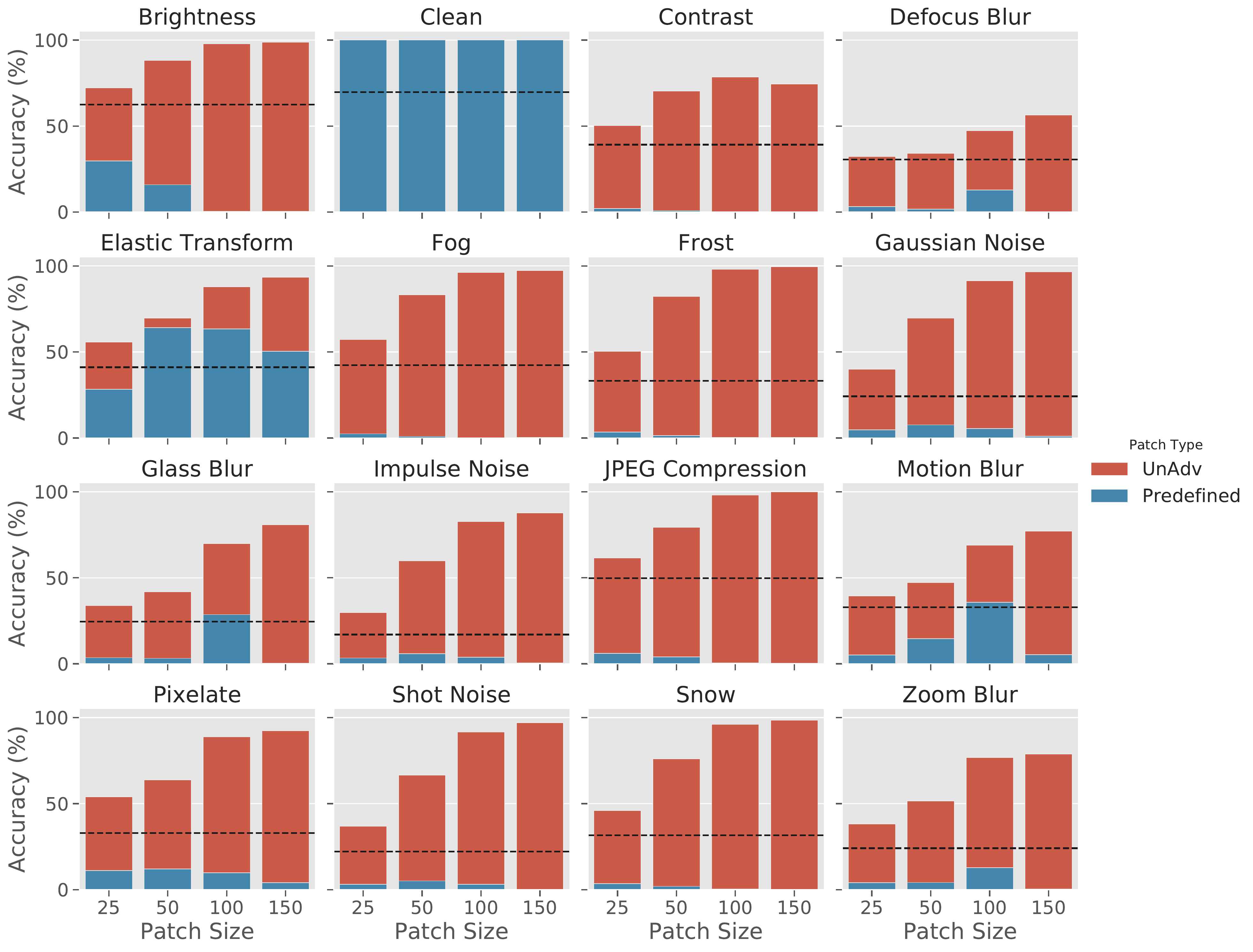}        
    \caption{Robustness of an ImageNet ResNet-18 model trained on a predefined patch.} 
    \label{fig:imagenet-predefined-patch}
\end{figure}

\begin{figure}[!htbp]
    \centering
    \includegraphics[width=.6\linewidth]{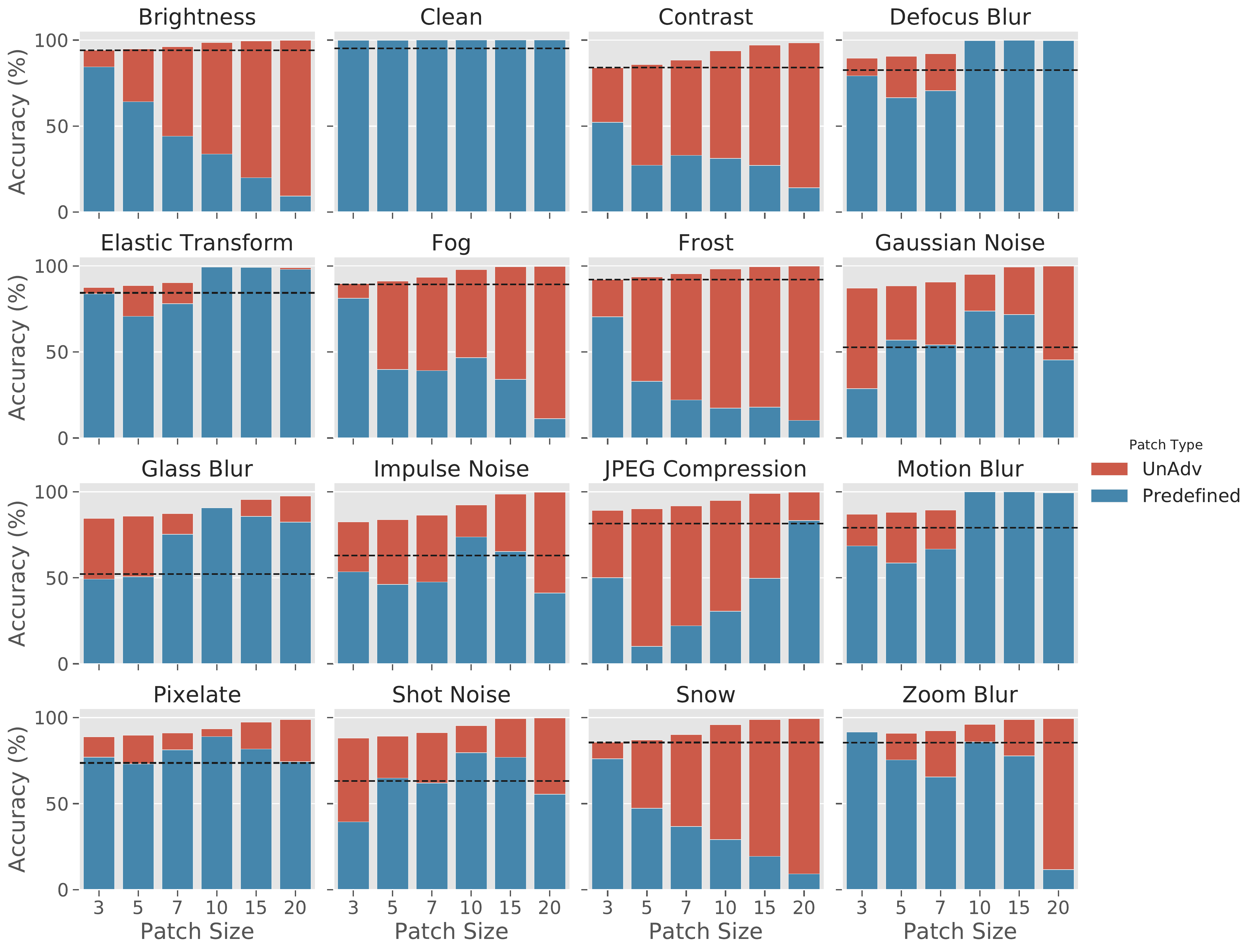}        
    \caption{Robustness of a CIFAR-10 ResNet-50 model trained on a predefined patch.} 
    \label{fig:cifar-predefined-patch}
\end{figure}

This baselines is slightly different than the previous baselines since it allows the underlying classification model to be changed. Basically, we fix the a set of patches to predefined pattern (here a fixed random gaussian noise for each class), and we train a classifier on an undversarial/boosted dataset with these patches. The resulting models are consistently weaker on all corruptions of ImageNet-C and  CFAR-10-C as shown in \autoref{fig:imagenet-predefined-patch} and \autoref{fig:cifar-predefined-patch} compared to our trained patches the main paper in \autoref{fig:2d_results}.

